\pdfoutput=1
\documentclass[11pt]{article}

\usepackage[table,xcdraw]{xcolor}
\usepackage{tabularx,booktabs,enumitem}

\usepackage{graphicx}
\usepackage{multirow}
\usepackage{amsmath}
\usepackage{caption}
\usepackage{textcomp}

\usepackage[final]{acl}

\usepackage{times}
\usepackage{latexsym}

\usepackage[T1]{fontenc}

\usepackage[utf8]{inputenc}

\usepackage{microtype}

\usepackage{inconsolata}

\title{Diverge to Induce Prompting: Multi-Rationale Induction for Zero-Shot Reasoning}

\author{
    Po-Chun Chen\textsuperscript{1} \quad    
    Hen-Hsen Huang\textsuperscript{2}\quad
    Hsin-Hsi Chen\textsuperscript{1,3}
    \\
    \textsuperscript{1}Department of Computer Science and Information Engineering, 
    \\ National Taiwan University, Taiwan
    \\
    \textsuperscript{2}Institute of Information Science, Academia Sinica, Taiwan
    \\
    \textsuperscript{3}AI Research Center (AINTU), National Taiwan University, Taiwan
    \\
    \texttt{pcchen@nlg.csie.ntu.edu.tw,}
    \\
    \texttt{hhhuang@iis.sinica.edu.tw, \quad hhchen@ntu.edu.tw}
}

\begin{document}
\maketitle

\begin{abstract}
To address the instability of unguided reasoning paths in standard Chain-of-Thought prompting, recent methods guide large language models (LLMs) by first eliciting a single reasoning strategy. However, relying on just one strategy for each question can still limit performance across diverse tasks. We propose Diverge-to-Induce Prompting (DIP), a framework that first prompts an LLM to generate multiple diverse high-level rationales for each question. Each rationale is then elaborated into a detailed, step-by-step draft plan. Finally, these draft plans are induced into a final plan. DIP enhances zero-shot reasoning accuracy without reliance on resource-intensive sampling. Experiments show that DIP outperforms single-strategy prompting, demonstrating the effectiveness of multi-plan induction for prompt-based reasoning.
\end{abstract}

\section{Introduction}

Prompt-based reasoning has become a central paradigm for eliciting logical behavior in large language models (LLMs), especially in zero-shot settings \citep{Wei2022ChainOT, Kojima2022LargeLM}. A common approach is zero-shot chain-of-thought (CoT) prompting, which guides models to produce reasoning chains directly but leaves the reasoning process unguided.

To address this limitation, recent work proposes single-strategy prompting methods, such as Plan-and-Solve prompting \citep{wang-etal-2023-plan} and Strategic Chain-of-Thought (S-CoT; \citealt{Wang2024StrategicCG}), where the model first generates a high-level plan or strategy before producing the final answer. While efficient, these approaches typically commit to a single path per question, relying on the model's first intuition and potentially missing better alternatives. Other methods generate multiple reasoning paths and select among them, for example through external voting, reranking, or sampling-based aggregation \citep{Wang2022SelfConsistencyIC, Zheng2023TakeAS, Suzgun2024MetaPromptingEL}, but these strategies require repeated model calls or additional selection modules, resulting in substantial computational overhead.

We introduce \textbf{Diverge-to-Induce Prompting (DIP)}, a prompting framework that first generates multiple high-level rationales, elaborates each into a step-by-step draft plan, and then induces a final plan from these draft plans, which is then used to perform the final inference. Inspired by advances in instruction induction \citep{honovich-etal-2023-instruction, chen-etal-2024-induct}, DIP enables LLMs to synthesize a high-quality, instance-level draft plan by integrating diverse perspectives. Experiments show that DIP outperforms strong baselines, including state-of-the-art single-strategy prompting methods, in most evaluation settings.

In summary, our main contributions are threefold: (1) We propose a multi-rationale induction framework that elicits multiple high-level rationales per question, enabling LLMs to integrate diverse perspectives and avoid the blind spots of single-path reasoning; (2) Our method induces a final plan from these draft plans using multiple rationales, without relying on repeated sampling, voting, or external selection modules; and (3) Experiments on BBH and LiveBench Reasoning tasks, spanning a wide range of LLM families, demonstrate that our approach outperforms state-of-the-art single-path reasoning methods in reasoning accuracy.

\begin{figure*}[t]
  \centering
  \includegraphics[width=\linewidth]{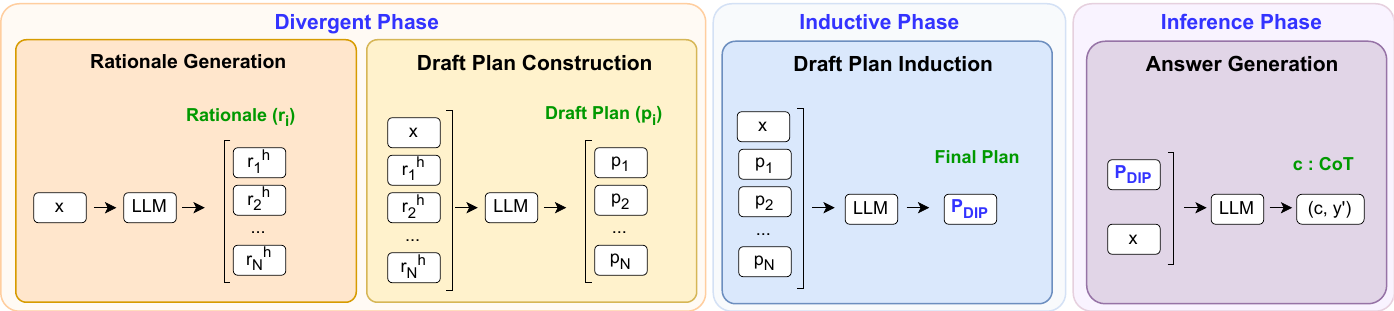}

    \caption{
      Overview of the DIP framework. The process consists of three main phases:
      a \textbf{\textcolor[rgb]{0.85,0.44,0.02}{Divergent Phase}}, where the model generates multiple high-level rationales and constructs a draft plan for each;
      an \textbf{\textcolor[rgb]{0.18,0.37,0.60}{Inductive Phase}}, where all plans are induced into a final plan;
      and an \textbf{\textcolor[rgb]{0.46,0.27,0.53}{Inference Phase}}, which produces the final reasoning and answer.
    }
  \label{fig:framework}
\end{figure*}

\section{DIP Framework}

DIP comprises three main phases: (1) a Divergent Phase, where the model generates multiple high-level rationales and constructs a draft plan for each; (2) an inductive phase, which consolidates these draft plans into a final plan; and (3) an Inference Phase, which produces the final reasoning and answer. Figure~\ref{fig:framework} illustrates the process, with complete prompt templates provided in Appendix Figures~\ref{tab:DIP_Prompt_template_1}--\ref{tab:DIP_Prompt_template_4}, which cover templates for rationale generation, draft plan construction, induction, and answer generation.

\subsection{Rationale Generation}
Given a question $x$, the model generates a set of $N$ high-level rationales in a single call:
\[
R = \{ r_1, r_2, \dots, r_N \} = \mathrm{LLM}(x)
\]
Each $r_i$ denotes a distinct high-level rationale for solving the question.

\subsection{Draft Plan Construction}
Each rationale $r_i$ is expanded by the LLM into a corresponding step-by-step draft plan $p_i$. All plans are generated in a single model call by providing $x$ and the set $R$ as input, resulting in $P = \{ p_1, p_2, \dots, p_N \}$, where each $p_i$ corresponds to $r_i$.
\[
P = \{ p_1, p_2, \dots, p_N \} = \mathrm{LLM}(x, R)
\]

\subsection{Draft Plan Induction}
The induction phase induces a single final plan that incorporates the diverse perspectives found in the set of draft plans $P$ for the same question $x$:
\[
P_{\text{DIP}} = \mathrm{LLM}(x, P)
\]

\subsection{Final Inference}
The induced draft plan $P_{\text{DIP}}$ is then used to answer the original question:
\[
(c, y^*) = \mathrm{LLM}(P_{\text{DIP}}, x)
\]
where $c$ denotes the chain-of-thought reasoning and $y^*$ is the final predicted answer.

\begin{table*}[ht]
  \small
  \centering
  \begin{tabular}{l|ccccc|ccccc|}
    \toprule
    \textbf{Model} & \multicolumn{5}{c}{\textbf{BBH}} & \multicolumn{5}{c}{\textbf{LiveBench Reasoning}} \\
     & \textbf{Z-CoT} & \textbf{R-CoT} & \textbf{S-CoT} & \textbf{Ours} & \textbf{$\Delta$} & \textbf{Z-CoT} & \textbf{R-CoT} & \textbf{S-CoT} & \textbf{Ours} & \textbf{$\Delta$} \\
    \midrule
    Llama 3.3 8B & 69.33 & 61.62 & 64.35 & \textbf{71.19} & \textcolor{blue}{+1.86} & 21.50 & 16.50 & 18.50 & \textbf{28.00} & \textcolor{blue}{+6.50} \\
    Llama 3.3 70B & 83.54 & 77.45 & 80.17 & \textbf{85.51} & \textcolor{blue}{+1.97} & 42.00 & 22.50 & 44.00 & \textbf{47.00} & \textcolor{blue}{+5.00} \\
    Llama 4 Scout & 77.74 & 79.25 & 70.32 & \textbf{84.46} & \textcolor{blue}{+6.72} & 12.50 & 8.00 & 13.50 & \textbf{43.00} & \textcolor{blue}{+30.50} \\
    Llama 4 Maverick & 83.59 & 84.64 & 78.67 & \textbf{86.20} & \textcolor{blue}{+2.61} & 55.00 & 49.50 & 50.00 & \textbf{56.50} & \textcolor{blue}{+1.50} \\
    \midrule
    Mistral Small 2 & 71.07 & 68.70 & 69.51 & \textbf{71.65} & \textcolor{blue}{+0.58} & 29.50 & 30.50 & 28.00 & \textbf{31.00} & \textcolor{blue}{+1.50} \\
    Mistral Small 3 & 78.78 & 77.33 & 77.97 & \textbf{81.04} & \textcolor{blue}{+2.26} & \textbf{36.50} & 36.00 & 35.50 & \textbf{36.50} & \textcolor{blue}{+0.00} \\
    Mistral Small 3.1 & 77.22 & 78.43 & 77.91 & \textbf{78.90} & \textcolor{blue}{+1.68} & 39.00 & 33.50 & 38.00 & \textbf{39.50} & \textcolor{blue}{+0.50} \\
    Mistral Medium 3 & 83.48 & 83.48 & 84.41 & \textbf{87.01} & \textcolor{blue}{+3.53} & 53.00 & 49.00 & 53.50 & \textbf{54.50} & \textcolor{blue}{+1.50} \\
    Mistral Large 2 & 84.06 & 82.49 & 82.20 & \textbf{84.87} & \textcolor{blue}{+0.81} & 44.50 & 42.00 & 43.50 & \textbf{46.50} & \textcolor{blue}{+2.00} \\

    \midrule
    Gemini 2.0 Flash Lite & 79.25 & 81.28 & 79.71 & \textbf{83.54} & \textcolor{blue}{+4.29} & 38.50 & 43.00 & 39.00 & \textbf{48.00} & \textcolor{blue}{+9.50} \\
    Gemini 2.0 Flash & 80.64 & 82.03 & 82.09 & \textbf{85.22} & \textcolor{blue}{+4.58} & 54.50 & 52.00 & 55.00 & \textbf{58.50} & \textcolor{blue}{+4.00} \\
    \midrule
    GPT 4o mini & 79.59 & 76.12 & 78.49 & \textbf{82.90} & \textcolor{blue}{+3.31} & 29.50 & 24.50 & 31.00 & \textbf{32.50} & \textcolor{blue}{+3.00} \\
    GPT 4o & 84.87 & 85.51 & 85.97 & \textbf{88.06} & \textcolor{blue}{+3.19} & 41.50 & \textbf{46.50} & 42.00 & 46.00 & \textcolor{blue}{+4.50} \\
    GPT 4.1 Nano & 76.17 & 74.90 & 76.64 & \textbf{77.91} & \textcolor{blue}{+1.74} & \textbf{35.00} & 29.00 & 29.00 & 32.00 & \textcolor{red}{-3.00} \\
    GPT 4.1 Mini & 88.93 & 89.51 & 88.93 & \textbf{89.86} & \textcolor{blue}{+0.93} & 52.00 & 54.50 & 63.00 & \textbf{65.00} & \textcolor{blue}{+13.00} \\
    GPT 4.1 & 89.04 & 88.87 & 90.09 & \textbf{92.35} & \textcolor{blue}{+3.31} & 65.50 & 64.50 & 63.00 & \textbf{70.50} & \textcolor{blue}{+5.00} \\
    o4 Mini & 89.68 & 91.48 & 90.96 & \textbf{91.59} & \textcolor{blue}{+1.91} & 82.50 & 80.00 & 84.00 & \textbf{91.00} & \textcolor{blue}{+8.50} \\
    \midrule
    Grok 3 & 89.33 & 87.54 & 89.04 & \textbf{90.61} & \textcolor{blue}{+1.28} & 75.00 & 72.00 & \textbf{77.00} & 73.00 & \textcolor{red}{-2.00} \\
    Grok 3 Mini (High) & 89.57 & 88.41 & 90.78 & \textbf{91.48} & \textcolor{blue}{+1.91} & 84.50 & 83.00 & 86.50 & \textbf{92.00} & \textcolor{blue}{+7.50} \\
    Grok 3 Mini (Low) & 89.04 & 86.38 & 89.91 & \textbf{90.26} & \textcolor{blue}{+1.22} & 82.00 & 81.00 & 84.00 & \textbf{84.50} & \textcolor{blue}{+2.50} \\
    \bottomrule
  \end{tabular}
  \caption{Zero-shot performance (\%) of different models under Z-CoT, R-CoT, S-CoT, our method, and their difference $\Delta$ (Ours $-$ Z-CoT). 
  Blue = improvement, Red = degradation.
  }
  \label{tab:livebench_zero_shot}
\end{table*}

\section{Experiments Setting}

\subsection{Models}

We evaluate DIP across a broad range of LLMs and Large Reasoning Models (LRMs), including both open-source and proprietary systems. Our experiments span six major families: \textbf{LLaMA}, \textbf{Mistral}, \textbf{Gemini}, \textbf{GPT}, \textbf{Grok}, and the \textbf{o-Series}, covering diverse model sizes and capabilities. In particular, our evaluation includes LRMs with extended reasoning capabilities (o4 Mini and Grok 3 Mini), which employ test-time compute for enhanced reasoning. Full model names, versions, and configuration details are provided in Appendix~\ref{sec:exp_config}.

\subsection{Datasets}
We use the BIG-Bench Hard (BBH) benchmark \cite{Suzgun2022ChallengingBT}, a suite of tasks curated from the BIG-Bench collection \cite{Srivastava2022BeyondTI} that current LLMs struggle to solve at average human performance. 
We also include the reasoning-type tasks from LiveBench \cite{livebench}, a challenging and contamination-controlled benchmark comprising diverse real-world problems, to evaluate models' high-precision reasoning abilities.

\subsection{Baselines}

\paragraph{Z-CoT} 
We apply Zero-shot Chain-of-Thought (Z-CoT) prompting \cite{Kojima2022LargeLM}, where the model is directly prompted to generate intermediate reasoning steps without demonstrations.

\paragraph{S-CoT} 
Automatic Strategic Chain-of-Thought (S-CoT), proposed by \citet{Wang2024StrategicCG}, is a state-of-the-art CoT prompting baseline. It first prompts the model to identify a high-level problem-solving strategy, which is then used to guide step-by-step reasoning.

\paragraph{R-CoT}
We design a variant of S-CoT called Rationale Chain-of-Thought (R-CoT), in which the model is prompted to generate a rationale instead of a general strategy before reasoning.

\subsection{Other Details}

In our experiments, we set \( N = 5 \), meaning that DIP generates five rationales per question. These are elaborated into draft plans and then induced into a final plan for inference. 

For model parameters, we set the temperature to 0 and Top-P to 1 whenever these options were supported by the model, ensuring deterministic and reproducible results. Additional implementation and evaluation details are provided in Appendix~\ref{sec:exp_config}.

\section{Results and Analysis}

We evaluate DIP on BBH and LiveBench Reasoning tasks using both LLMs and LRMs across multiple model families (Table~\ref{tab:livebench_zero_shot}). 
DIP outperforms all baselines, achieving the highest accuracy across all models on BBH, and obtaining the best results in 17 out of 20 settings on LiveBench, demonstrating strong generalization.
The performance gains of DIP exhibit different patterns across benchmarks. On BBH, improvements over Z-CoT are consistent, ranging from 0.58 to 6.72 in accuracy. On the more challenging LiveBench tasks, DIP improves performance for most models, with gains ranging from 0.5 to 30.50. In particular, Llama 4 Scout and GPT 4.1 Mini achieve notable improvements of 30.50 and 13.00, respectively.

Notably, our approach surpasses both R-CoT and S-CoT in the vast majority of settings. Unlike these baselines, which generate only a single draft plan per question, DIP first explores multiple diverse rationales, constructs corresponding step-by-step plans, and then induces a final plan through draft plan induction. This multi-stage process enables improved accuracy reasoning across model types.
Moreover, for complex tasks, DIP helps models such as \texttt{Llama 4 Scout} better follow output format requirements, addressing a common failure case under baseline prompting and further contributing to improved accuracy.

\begin{table}[t!]
  \small
  \centering
  \begin{tabular}{l|ccc}
    \toprule
    \textbf{Model} & \textbf{DIP} & \textbf{DIP-R} & $\boldsymbol{\Delta}$ \\
    \midrule
    Llama 4 Scout & \textbf{84.46} & 80.23 & \textcolor{blue}{+4.23} \\
    Llama 4 Maverick & \textbf{86.20} & 86.03 & \textcolor{blue}{+0.17} \\
    \midrule
    Mistral Small 3 & \textbf{81.04} & 79.65 & \textcolor{blue}{+1.39} \\
    Mistral Medium 3 & \textbf{87.01} & 85.39 & \textcolor{blue}{+1.62} \\
    \midrule
    Gemini 2.0 Flash Lite & 83.42 & \textbf{84.00} & \textcolor{red}{-0.58} \\
    Gemini 2.0 Flash & \textbf{85.22} & 83.36 & \textcolor{blue}{+1.86} \\
    \midrule
    GPT 4.1 Mini & \textbf{89.86} & 89.39 & \textcolor{blue}{+0.47} \\
    O4 Mini & \textbf{91.59} & 90.84 & \textcolor{blue}{+0.75} \\
    \midrule
    Grok 3 & 90.61 & 90.61 & 0.00 \\
    Grok 3 Mini (Low) & \textbf{90.26} & 89.91 & \textcolor{blue}{+0.35} \\
    \bottomrule
  \end{tabular}
  \caption{Performance (\%) on BBH. $\Delta =$ DIP $-$ DIP-R. Blue = improvement, Red = degradation.}
  \label{tab:bbh_ours_ablation}
\end{table}

\begin{table}[ht]
  \small
  \setlength{\tabcolsep}{5.7pt}
  \centering
  \begin{tabular}{l|cccc}
    \toprule
    \textbf{Model} & \multicolumn{4}{c}{\textbf{BBH}} \\
     & \textbf{N=1} & \textbf{N=3} & \textbf{N=5} & \textbf{N=7} \\
    \midrule
    Llama 4 Scout & 83.65 & 84.12 & \textbf{84.46} & 82.38 \\
    Llama 4 Maverick & 86.96 & 86.20 & 86.20 & \textbf{87.77} \\
    \midrule
    Mistral Small 3 & 80.06 & 78.49 & 81.04 & \textbf{81.97} \\
    Mistral Medium 3 & 86.78 & 87.88 & 87.01 & \textbf{88.17} \\
    \midrule
    Gemini 2.0 Flash Lite & 81.04 & 83.71 & 83.42 & \textbf{84.17} \\
    Gemini 2.0 Flash & 85.10 & 83.65 & \textbf{85.22} & 82.55 \\
    \midrule
    GPT 4.1 Mini & 89.68 & 89.91 & 89.86 & \textbf{90.09} \\
    O4 Mini & 90.49 & 90.72 & \textbf{91.59} & 90.03 \\
    \midrule
    Grok 3 & 90.61 & 90.78 & 90.61 & \textbf{91.13} \\
    Grok 3 Mini (Low) & 89.39 & \textbf{91.83} & 90.26 & 91.13 \\
    \midrule
    Win N=1 Rate & - & 7/10 & 8/10 & 7/10 \\
    \bottomrule
  \end{tabular}
  \caption{Performance (\%) of different models on \textsc{BBH} as a function of the number of rationales $N$. For each $N>1$, the last row reports the number of models that outperform $N=1$.}
  \label{tab:nshot_bbh}
\end{table}

\section{Discussion}

Due to cost constraints, all ablation studies in this section are conducted on two models per series.

\paragraph{Effect of Rationale Generation}

To assess the impact of rationale generation, we compare DIP with an ablation variant, DIP-R, which omits the Rationale Generation step and directly constructs multiple draft plans for induction. As shown in Table~\ref{tab:bbh_ours_ablation}, DIP outperforms DIP-R on 9 out of 10 models. This confirms that prompting the model to first generate diverse rationales yields more robust and accurate plans.

\paragraph{Impact of Draft Plan Number}

We further analyze the effect of varying the number of draft plans ($N$) used in the induction step. As shown in Table~\ref{tab:nshot_bbh}, using multiple plans ($N>1$) outperforms $N=1$ in most settings. While the optimal $N$ varies across models and datasets, values between 5 and 7 generally yield the best results. This suggests that moderate diversity in draft plans provides sufficient signal for effective induction.
However, several models show performance degradation at higher $N$ values (e.g., Llama 4 Scout drops from 84.46\% at N=5 to 82.38\% at N=7), suggesting that excessive rationales may introduce noise that interferes with the induction process.

\begin{figure}[t!]
  \centering
  \small
  \renewcommand{\arraystretch}{1.1}
  \begin{tabular}{p{7.2cm}}
    \toprule
    \textbf{\textcolor{blue}{Source}} \\
    Die Mexikanische Königsnatter oder San-Luis-Potosi-Königsnatter ist eine Schlange aus der Familie der Nattern. \\
    \midrule
    \textbf{\textcolor{blue}{Translation}} \\
    The Mexican King's Snake or San Luis Potosi King snake is a snake of the Lantern family. \\
    \midrule
    \textbf{\textcolor{blue}{Options}} \\
    (A) Modifiers or Adjectives,\ (B) Numerical Values,\ (C) Negation or Antonyms,\ (D) Named Entities,\ (E) Dropped Content,\ (F) Facts \\
    \midrule
    \textbf{\textcolor{blue}{Answer}} \\
    (D) Named Entities \\
    \bottomrule
  \end{tabular}
  \vspace{-0.7em}
    \caption{\small 
    BBH Salient Translation Error Detection example. Identify the error type in a German-to-English translation. Here, ``Nattern'' (colubrid family) is mistranslated as ``Lantern family.''
    }
    
  \label{fig:case-study-example}
\end{figure}

\paragraph{Case Study}
Figure~\ref{fig:case-study-example} presents a BBH Salient Translation Error Detection example with the Llama 4 Scout model. In this instance, “Nattern” (colubrid family) is mistranslated as “Lantern family,” constituting a named entity error. Only our full DIP setting ($N=5$) induces a detailed plan that aligns family names and systematically checks all error types, leading to the correct answer. In contrast, both $N=1$ and DIP-R produce generic or incomplete plans and fail on this case. All other baselines also fail to identify the correct error type. 
Full intermediate outputs and prompts are provided in the Appendix (see Figures~\ref{sec:full-case-Reasoning-Approaches} to \ref{tab:RCOTTranslationError}).

\begin{table*}[t]
\small
\centering
\begin{tabular}{l|cccc|cccc}
\toprule
\multirow{3}{*}{\textbf{Method}} & \multicolumn{4}{c|}{\textbf{Llama 4 Scout (109B)}} & \multicolumn{4}{c}{\textbf{Llama 4 Maverick (400B)}} \\
\cmidrule(lr){2-5} \cmidrule(lr){6-9}
 & & \multicolumn{3}{c|}{\textbf{Tokens}} & & \multicolumn{3}{c}{\textbf{Tokens}} \\
\cmidrule(lr){3-5} \cmidrule(lr){7-9}
 & \textbf{Acc (\%)} & \textbf{Input} & \textbf{Output} & \textbf{Total} & \textbf{Acc (\%)} & \textbf{Input} & \textbf{Output} & \textbf{Total} \\
\midrule
Z-CoT & 77.74 & 298.54 & 376.63 & 675.17 & 83.59 & 298.54 & 409.84 & 708.38 \\
S-CoT & 70.32 & 333.54 & 426.89 & 760.43 & 78.67 & 333.09 & 530.61 & 863.70 \\
DIP & \textbf{84.46} & 705.08 & 1,555.51 & 2,260.59 & \textbf{86.20} & 704.40 & 1,480.99 & 2,185.39 \\
\midrule
Z-CoT+SC(k=3) & 82.55 & 895.62 & 1,129.89 & 2,025.51 & 85.04 & 895.62 & 1,229.52 & 2,125.14 \\
Z-CoT+SC(k=5) & 84.41 & 1,492.70 & 1,883.15 & 3,375.85 & 85.51 & 1,492.70 & 2,049.20 & 3,541.90 \\
Z-CoT+SC(k=10) & 84.17 & 2,985.40 & 3,766.30 & 6,751.70 & 85.51 & 2,985.40 & 4,098.40 & 7,083.80 \\
Z-CoT+SC(k=15) & 84.06 & 4,478.10 & 5,649.45 & 10,127.55 & 85.22 & 4,478.10 & 6,147.60 & 10,625.70 \\
Z-CoT+SC(k=20) & 84.17 & 5,970.80 & 7,532.60 & 13,503.40 & 86.09 & 5,970.80 & 8,196.80 & 14,167.60 \\
\midrule
S-CoT+SC(k=3) & 76.64 & 1,000.62 & 1,280.67 & 2,281.29 & 84.46 & 999.27 & 1,591.83 & 2,591.10 \\
S-CoT+SC(k=5) & 79.83 & 1,667.70 & 2,134.45 & 3,802.15 & 85.16 & 1,665.45 & 2,653.05 & 4,318.50 \\
S-CoT+SC(k=10) & 82.72 & 3,335.40 & 4,268.90 & 7,604.30 & 86.09 & 3,330.90 & 5,306.10 & 8,637.00 \\
S-CoT+SC(k=15) & 82.90 & 5,003.10 & 6,403.35 & 11,406.45 & 85.91 & 4,996.35 & 7,959.15 & 12,955.50 \\
S-CoT+SC(k=20) & 83.01 & 6,670.80 & 8,537.80 & 15,208.60 & 85.68 & 6,661.80 & 10,612.20 & 17,274.00 \\
\midrule
DIP+SC(k=3) & 84.87 & 2,015.02 & 2,242.73 & 4,257.75 & 86.38 & 1,861.18 & 2,284.95 & 4,146.13 \\
DIP+SC(k=5) & 84.81 & 3,515.50 & 2,929.95 & 6,445.45 & 86.38 & 3,209.18 & 3,088.91 & 6,298.09 \\
DIP+SC(k=10) & 84.58 & 7,387.43 & 4,648.00 & 12,035.43 & \textbf{86.90} & 6,698.21 & 5,098.81 & 11,797.02 \\
DIP+SC(k=15) & \textbf{84.99} & 12,154.98 & 6,366.05 & 18,521.03 & 86.38 & 11,082.86 & 7,108.71 & 18,191.57 \\
DIP+SC(k=20) & 84.64 & 16,922.53 & 8,084.10 & 25,006.63 & 86.55 & 15,467.51 & 9,118.61 & 24,586.12 \\
\bottomrule
\end{tabular}
\caption{Cost-performance comparison on BBH. Input tokens measure prompt cost, output tokens measure generation cost. For DIP+SC methods, SC is applied only to the final answer generation step.}
\label{tab:sc_comparison}
\end{table*}

\paragraph{Computational Cost Analysis}
To evaluate whether DIP's improvements stem from effective plan induction or simply increased computation, we compare against Self-Consistency (SC)~\citep{Wang2022SelfConsistencyIC}, a widely-used multi-path baseline. Due to cost constraints, we conduct this analysis on Llama 4 Scout and Maverick, the latest open-source Llama models. Table~\ref{tab:sc_comparison} reports token consumption, focusing on output tokens since they typically cost approximately 3× more than input tokens~\citep{together2025pricing}.

DIP achieves substantially better cost-performance trade-offs than sampling baselines. We primarily compare against SC with k=20, as prior work shows performance gains typically saturate beyond this point~\citep{Wang2022SelfConsistencyIC}. On Llama 4 Scout, DIP reaches 84.46\% with 1,556 output tokens versus 84.17\% with 7,533 tokens for Z-CoT+SC(k=20) (4.8× fewer tokens). Similar efficiency gains hold on Llama 4 Maverick: DIP uses 7.2× fewer tokens than S-CoT+SC(k=20) (1,481 vs. 10,612) while achieving higher accuracy (86.20\% vs. 85.68\%).

Even accounting for total token consumption including DIP's multi-stage prompting, DIP remains substantially more efficient. For example, on Llama 4 Maverick, DIP uses 2,185 total tokens versus 14,168 for Z-CoT+SC(k=20) (6.5× fewer tokens). Moreover, by applying SC only to the final answer generation step, DIP+SC(k=10) achieves our highest accuracy of 86.90\% with 11,797 total tokens, demonstrating that multi-rationale induction and sampling are complementary while maintaining competitive efficiency.

\section{Related Work}

\paragraph{Inductive Capacities of LLMs}
Recent studies affirm that LLMs possess strong inductive reasoning abilities, allowing them to abstract rules or instructions from limited examples and generalize to new instances~\citep{Wang2023HypothesisSI,Zhu2023LargeLM,Cheng2024InductiveOD,Sun2024ItDLL,He2024IDEAET,Yang2022LanguageMA,Honovich2022InstructionIF,chen-etal-2024-induct,10.1145/3746252.3760963}. Most prior work focuses on rule induction at the dataset or task level. By contrast, our work explores \textbf{instance-level induction}: for each question, we prompt the model to induce a final solution from multiple draft plans, leveraging LLMs’ inductive capacity at the instance level.

\paragraph{Plan Generation by LLMs}
LLMs have also been used to generate draft plans before problem solving. Plan-and-Solve and Strategic Chain-of-Thought (S-CoT) prompt the model to produce a high-level plan or strategy to guide step-by-step reasoning~\citep{wang-etal-2023-plan,Wang2024StrategicCG}. Search-based methods such as Tree-of-Thoughts (ToT), Reasoning via Planning, and Graph-of-Thoughts expand multiple reasoning paths~\citep{10.5555/3666122.3666639,hao-etal-2023-reasoning,10.1609/aaai.v38i16.29720}. 

These search-based approaches, exemplified by ToT, adopt a ``learning by doing'' paradigm: they immediately execute reasoning steps, explore multiple paths in parallel, and iteratively evaluate partial solutions to guide further exploration. In contrast, DIP follows a ``planning before doing'' approach. We first generate multiple diverse abstract planning strategies without execution, synthesize them into a final plan through induction, and only then perform a single execution. This approach achieves both the diversity of multi-plan exploration and the efficiency of single-pass inference.

\section{Conclusion}

We present DIP, a prompting framework that enhances zero-shot reasoning in large language models by first generating multiple diverse high-level rationales, elaborating a draft plan for each, and then inducing a final plan from these. Experiments on BBH and LiveBench Reasoning tasks show that DIP outperforms strong baselines, including state-of-the-art single-strategy prompting methods, across diverse models and tasks. These results demonstrate that multi-rationale induction offers a practical and effective approach for more robust prompt-based reasoning at the instance level.

\section{Limitations}

While DIP demonstrates improvements on BBH and LiveBench Reasoning tasks, it has several limitations. First, generating and processing multiple rationales and plans per question leads to higher computational cost and more API calls than basic prompting baselines such as Z-CoT. This overhead may limit applicability in cost- or latency-sensitive scenarios, including large-scale or real-time applications.
Second, DIP's performance gains are mainly observed on tasks requiring complex, multi-step reasoning. For tasks that involve only straightforward fact retrieval or commonsense question answering, the framework may offer limited or no benefit, as generating diverse rationales can introduce unnecessary complexity.
Finally, our evaluation is restricted to English benchmarks and a subset of LLM architectures. The generalizability of DIP to multilingual tasks, domain-specific reasoning, or other model families remains to be validated.

\section*{Acknowledgements}
This work was supported by National Science and Technology Council, Taiwan, under grants NSTC 113-2634-F-002-003- and 114-2221-E-002-070-MY3, and Ministry of Education (MOE), Taiwan, under grants NTU-113L9009, NTU-114L9009, and NTU-114L900901.

\bibliography{custom}

\begin{thebibliography}{38}
\providecommand{\natexlab}[1]{#1}

\bibitem[{Besta et~al.(2024)Besta, Blach, Kubicek, Gerstenberger, Podstawski, Gianinazzi, Gajda, Lehmann, Niewiadomski, Nyczyk, and Hoefler}]{10.1609/aaai.v38i16.29720}
Maciej Besta, Nils Blach, Ales Kubicek, Robert Gerstenberger, Micha\l{} Podstawski, Lukas Gianinazzi, Joanna Gajda, Tomasz Lehmann, Hubert Niewiadomski, Piotr Nyczyk, and Torsten Hoefler. 2024.
\newblock \href {https://doi.org/10.1609/aaai.v38i16.29720} {Graph of thoughts: solving elaborate problems with large language models}.
\newblock In \emph{Proceedings of the Thirty-Eighth AAAI Conference on Artificial Intelligence and Thirty-Sixth Conference on Innovative Applications of Artificial Intelligence and Fourteenth Symposium on Educational Advances in Artificial Intelligence}, AAAI'24/IAAI'24/EAAI'24. AAAI Press.

\bibitem[{Chen et~al.(2025)Chen, Huang, and Chen}]{10.1145/3746252.3760963}
Po-Chun Chen, Hen-Hsen Huang, and Hsin-Hsi Chen. 2025.
\newblock \href {https://doi.org/10.1145/3746252.3760963} {Vqa-induct: Instruction induction for visual question answering}.
\newblock In \emph{Proceedings of the 34th ACM International Conference on Information and Knowledge Management}, CIKM '25, page 4659–4664, New York, NY, USA. Association for Computing Machinery.

\bibitem[{Chen et~al.(2024)Chen, Wei, Huang, and Chen}]{chen-etal-2024-induct}
Po-Chun Chen, Sheng-Lun Wei, Hen-Hsen Huang, and Hsin-Hsi Chen. 2024.
\newblock \href {https://doi.org/10.18653/v1/2024.emnlp-main.297} {Induct-learn: Short phrase prompting with instruction induction}.
\newblock In \emph{Proceedings of the 2024 Conference on Empirical Methods in Natural Language Processing}, pages 5204--5231, Miami, Florida, USA. Association for Computational Linguistics.

\bibitem[{Cheng et~al.(2024)Cheng, Yang, Jiang, Wang, Huang, Li, Li, Li, Gao, Li, Yin, and Sun}]{Cheng2024InductiveOD}
Kewei Cheng, Jingfeng Yang, Haoming Jiang, Zhengyang Wang, Binxuan Huang, Ruirui Li, Shiyang Li, Zheng Li, Yifan Gao, Xian Li, Bing Yin, and Yizhou Sun. 2024.
\newblock \href {https://api.semanticscholar.org/CorpusID:271600705} {Inductive or deductive? rethinking the fundamental reasoning abilities of llms}.
\newblock \emph{ArXiv}, abs/2408.00114.

\bibitem[{Dubey et~al.(2024)Dubey, Jauhri, Pandey, Kadian, Al-Dahle, Letman, Mathur, Schelten, Yang, Fan, Goyal, Hartshorn, Yang, Mitra, Sravankumar, Korenev, Hinsvark, Rao, Zhang, Rodriguez, Gregerson, Spataru, Rozi{\`e}re, Biron, Tang, Chern, Caucheteux, Nayak, Bi, Marra, McConnell, Keller, Touret, Wu, Wong, tian Cant{\'o}n~Ferrer, Nikolaidis, Allonsius, Song, Pintz, Livshits, Esiobu, Choudhary, Mahajan, Garcia-Olano, Perino, Hupkes, Lakomkin, AlBadawy, Lobanova, Dinan, Smith, Radenovic, Zhang, Synnaeve, Lee, Anderson, Nail, Mialon, Pang, Cucurell, Nguyen, Korevaar, Xu, Touvron, Zarov, Ibarra, Kloumann, Misra, Evtimov, Copet, Lee, Geffert, Vranes, Park, Mahadeokar, Shah, van~der Linde, Billock, Hong, Lee, Fu, Chi, Huang, Liu, Wang, Yu, Bitton, Spisak, Park, Rocca, Johnstun, Saxe, Jia, Alwala, Upasani, Plawiak, Li, neth Heafield, Stone, El-Arini, Iyer, Malik, ley Chiu, Bhalla, Rantala-Yeary, van~der Maaten, Chen, Tan, Jenkins, Martin, Madaan, Malo, Blecher, Landzaat, de~Oliveira, Muzzi, Pasupuleti, Singh,
  Paluri, Kardas, Oldham, Rita, Pavlova, Kambadur, Lewis, Si, Singh, Hassan, Goyal, Torabi, lay Bashlykov, Bogoychev, Chatterji, Duchenne, cCelebi, Alrassy, Zhang, Li, Vasi{\'c}, Weng, Bhargava, Dubal, Krishnan, Koura, Xu, He, Dong, Srinivasan, Ganapathy, Calderer, Cabral, Stojnic, Raileanu, Girdhar, Patel, main Sauvestre, nie Polidoro, Sumbaly, Taylor, Silva, Hou, Wang, Hosseini, hana Chennabasappa, Singh, Bell, Kim, Edunov, Nie, Narang, Raparthy, Shen, Wan, Bhosale, Zhang, Vandenhende, Batra, Whitman, Sootla, Collot, Gururangan, Borodinsky, Herman, Fowler, Sheasha, Georgiou, Scialom, Speckbacher, Mihaylov, Xiao, Karn, Goswami, Gupta, Ramanathan, Kerkez, Gonguet, ginie Do, Vogeti, Petrovic, Chu, Xiong, Fu, ney Meers, Martinet, Wang, Tan, Xie, Jia, Wang, Goldschlag, Gaur, Babaei, Wen, Song, Zhang, Li, Mao, Coudert, Yan, Chen, Papakipos, Singh, Grattafiori, Jain, Kelsey, Shajnfeld, Gangidi, Victoria, Goldstand, Menon, Sharma, Boesenberg, Vaughan, Baevski, Feinstein, Kallet, Sangani, Yunus, Lupu, Alvarado,
  Caples, Gu, Ho, Poulton, Ryan, Ramchandani, Franco, Saraf, Chowdhury, Gabriel, Bharambe, Eisenman, Yazdan, James, Maurer, Leonhardi, Huang, Loyd, de~Paola, Paranjape, Liu, Wu, Ni, Hancock, Wasti, Spence, Stojkovic, Gamido, Montalvo, Parker, Burton, Mejia, Wang, Kim, Zhou, Hu, Chu, Cai, Tindal, Feichtenhofer, Civin, Beaty, Kreymer, Li, Wyatt, Adkins, Xu, Testuggine, David, Parikh, Liskovich, Foss, Wang, Le, Holland, Dowling, Jamil, Montgomery, Presani, Hahn, Wood, Brinkman, Arcaute, Dunbar, Smothers, Sun, Kreuk, Tian, Ozgenel, Caggioni, Guzm’an, Kanayet, Seide, Florez, Schwarz, Badeer, Swee, Halpern, Thattai, Herman, Sizov, Zhang, Lakshminarayanan, Shojanazeri, Zou, Wang, Zha, Habeeb, Rudolph, Suk, Aspegren, Goldman, Molybog, Tufanov, Veliche, Gat, Weissman, Geboski, Kohli, Asher, Gaya, Marcus, Tang, Chan, Zhen, Reizenstein, Teboul, Zhong, Jin, Yang, Cummings, Carvill, Shepard, McPhie, Torres, Ginsburg, Wang, Wu, KamHou, Saxena, Prasad, Khandelwal, Zand, Matosich, Veeraraghavan, Michelena, Li, Huang,
  Chawla, Lakhotia, Huang, Chen, Garg, Lavender, Silva, Bell, Zhang, Guo, Yu, Moshkovich, Wehrstedt, Khabsa, Avalani, Bhatt, Tsimpoukelli, Mankus, Hasson, Lennie, Reso, Groshev, Naumov, Lathi, Keneally, Seltzer, Valko, Restrepo, Patel, Vyatskov, Samvelyan, Clark, Macey, Wang, Hermoso, Metanat, Rastegari, Bansal, Santhanam, Parks, White, Bawa, Singhal, Egebo, Usunier, Laptev, Dong, Zhang, Cheng, Chernoguz, Hart, Salpekar, Kalinli, Kent, Parekh, Saab, Balaji, dro Rittner, Bontrager, Roux, Doll{\'a}r, Zvyagina, Ratanchandani, Yuvraj, Liang, Alao, Rodriguez, Ayub, Murthy, Nayani, Mitra, Li, Hogan, Battey, Wang, Maheswari, Howes, Rinott, Bondu, Datta, Chugh, Hunt, Dhillon, Sidorov, Pan, Verma, Yamamoto, Ramaswamy, Lindsay, Feng, Lin, Zha, Shankar, Zhang, Wang, Agarwal, Sajuyigbe, Chintala, Max, Chen, Kehoe, Satterfield, Govindaprasad, Gupta, Cho, Virk, Subramanian, Choudhury, Goldman, Remez, Glaser, Best, Kohler, Robinson, Li, Zhang, Matthews, Chou, Shaked, Vontimitta, Ajayi, Montanez, Mohan, Kumar, Mangla,
  Ionescu, Poenaru, Mihailescu, Ivanov, Li, Wang, Jiang, Bouaziz, Constable, Tang, Wang, Wu, Wang, Xia, Wu, Gao, Chen, Hu, Jia, Qi, Li, Zhang, Zhang, Adi, Nam, Wang, Hao, Qian, He, Rait, DeVito, Rosnbrick, Wen, Yang, and Zhao}]{Dubey2024TheL3}
Abhimanyu Dubey, Abhinav Jauhri, Abhinav Pandey, Abhishek Kadian, Ahmad Al-Dahle, Aiesha Letman, Akhil Mathur, Alan Schelten, Amy Yang, Angela Fan, Anirudh Goyal, Anthony~S. Hartshorn, Aobo Yang, Archi Mitra, Archie Sravankumar, Artem Korenev, Arthur Hinsvark, Arun Rao, Aston Zhang, and 510 others. 2024.
\newblock \href {https://api.semanticscholar.org/CorpusID:271571434} {The llama 3 herd of models}.
\newblock \emph{ArXiv}, abs/2407.21783.

\bibitem[{Google(2025)}]{google2025gemini2}
Google. 2025.
\newblock \href {https://developers.googleblog.com/en/gemini-2-family-expands/} {Gemini 2.0: Flash, flash-lite and pro}.
\newblock FEB. 5, 2025.

\bibitem[{Hao et~al.(2023)Hao, Gu, Ma, Hong, Wang, Wang, and Hu}]{hao-etal-2023-reasoning}
Shibo Hao, Yi~Gu, Haodi Ma, Joshua Hong, Zhen Wang, Daisy Wang, and Zhiting Hu. 2023.
\newblock \href {https://doi.org/10.18653/v1/2023.emnlp-main.507} {Reasoning with language model is planning with world model}.
\newblock In \emph{Proceedings of the 2023 Conference on Empirical Methods in Natural Language Processing}, pages 8154--8173, Singapore. Association for Computational Linguistics.

\bibitem[{He et~al.(2024)He, Zhang, Yan, Wu, and Chen}]{He2024IDEAET}
Kaiyu He, Mian Zhang, Shuo Yan, Peilin Wu, and Zhiyu~Zoey Chen. 2024.
\newblock \href {https://api.semanticscholar.org/CorpusID:273025550} {Idea: Enhancing the rule learning ability of large language model agent through induction, deduction, and abduction}.

\bibitem[{Honovich et~al.(2022)Honovich, Shaham, Bowman, and Levy}]{Honovich2022InstructionIF}
Or~Honovich, Uri Shaham, Samuel~R. Bowman, and Omer Levy. 2022.
\newblock \href {https://api.semanticscholar.org/CorpusID:248986755} {Instruction induction: From few examples to natural language task descriptions}.
\newblock \emph{ArXiv}, abs/2205.10782.

\bibitem[{Honovich et~al.(2023)Honovich, Shaham, Bowman, and Levy}]{honovich-etal-2023-instruction}
Or~Honovich, Uri Shaham, Samuel~R. Bowman, and Omer Levy. 2023.
\newblock \href {https://doi.org/10.18653/v1/2023.acl-long.108} {Instruction induction: From few examples to natural language task descriptions}.
\newblock In \emph{Proceedings of the 61st Annual Meeting of the Association for Computational Linguistics (Volume 1: Long Papers)}, pages 1935--1952, Toronto, Canada. Association for Computational Linguistics.

\bibitem[{Kojima et~al.(2022)Kojima, Gu, Reid, Matsuo, and Iwasawa}]{Kojima2022LargeLM}
Takeshi Kojima, Shixiang~Shane Gu, Machel Reid, Yutaka Matsuo, and Yusuke Iwasawa. 2022.
\newblock \href {https://api.semanticscholar.org/CorpusID:249017743} {Large language models are zero-shot reasoners}.
\newblock \emph{ArXiv}, abs/2205.11916.

\bibitem[{MetaAI(2025{\natexlab{a}})}]{meta_2025_llama3_3}
MetaAI. 2025{\natexlab{a}}.
\newblock \href {https://www.llama.com/products/llama-api/} {Introducing llama api}.
\newblock 2025.

\bibitem[{MetaAI(2025{\natexlab{b}})}]{meta_2025_llama4}
MetaAI. 2025{\natexlab{b}}.
\newblock \href {https://ai.meta.com/blog/llama-4-multimodal-intelligence/} {The llama 4 herd: The beginning of a new era of natively multimodal ai innovation}.
\newblock April 5, 2025.

\bibitem[{MistralAI(2024{\natexlab{a}})}]{mistral2024small2}
MistralAI. 2024{\natexlab{a}}.
\newblock \href {https://mistral.ai/news/september-24-release} {Ai in abundance}.
\newblock Sep 17, 2024.

\bibitem[{MistralAI(2024{\natexlab{b}})}]{mistral2024Pixtral_Large}
MistralAI. 2024{\natexlab{b}}.
\newblock \href {https://mistral.ai/news/pixtral-large} {Pixtral large}.
\newblock Nov 18, 2024.

\bibitem[{MistralAI(2025{\natexlab{a}})}]{mistral_2025_medium_3}
MistralAI. 2025{\natexlab{a}}.
\newblock \href {https://mistral.ai/news/mistral-medium-3} {Medium is the new large.}
\newblock May 7, 2025.

\bibitem[{MistralAI(2025{\natexlab{b}})}]{mistral2025small3}
MistralAI. 2025{\natexlab{b}}.
\newblock \href {https://mistral.ai/news/mistral-small-3} {Mistral small 3}.
\newblock Jan 30, 2025.

\bibitem[{MistralAI(2025{\natexlab{c}})}]{mistral2025small3_1}
MistralAI. 2025{\natexlab{c}}.
\newblock \href {https://mistral.ai/news/mistral-small-3-1} {Mistral small 3.1}.
\newblock Mar 17, 2025.

\bibitem[{OpenAI(2024{\natexlab{a}})}]{openai2024gpt4omini}
OpenAI. 2024{\natexlab{a}}.
\newblock \href {https://openai.com/index/gpt-4o-mini-advancing-cost-efficient-intelligence/} {Gpt-4o mini: advancing cost-efficient intelligence}.
\newblock July 18, 2024.

\bibitem[{OpenAI(2024{\natexlab{b}})}]{openai2024gpt4o}
OpenAI. 2024{\natexlab{b}}.
\newblock \href {https://openai.com/index/hello-gpt-4o/} {Hello gpt-4o}.
\newblock May 13, 2024.

\bibitem[{OpenAI(2025{\natexlab{a}})}]{openai_2025_GPT_4_1}
OpenAI. 2025{\natexlab{a}}.
\newblock \href {https://openai.com/index/gpt-4-1/} {Introducing gpt-4.1 in the api}.
\newblock April 14, 2025.

\bibitem[{OpenAI(2025{\natexlab{b}})}]{openai_2025_o4_mini}
OpenAI. 2025{\natexlab{b}}.
\newblock \href {https://openai.com/index/introducing-o3-and-o4-mini/} {Introducing openai o3 and o4-mini}.
\newblock April 16, 2025.

\bibitem[{Srivastava et~al.(2022)Srivastava, Rastogi, Rao, Shoeb, Abid, Fisch, Brown, Santoro, Gupta, Garriga-Alonso, Kluska, Lewkowycz, Agarwal, Power, Ray, Warstadt, Kocurek, Safaya, Tazarv, Xiang, Parrish, Nie, Hussain, Askell, Dsouza, Slone, Rahane, Iyer, Andreassen, Madotto, Santilli, Stuhlmuller, Dai, La, Lampinen, Zou, Jiang, Chen, Vuong, Gupta, Gottardi, Norelli, Venkatesh, Gholamidavoodi, Tabassum, Menezes, Kirubarajan, Mullokandov, Sabharwal, Herrick, Efrat, Erdem, Karakacs, Roberts, Loe, Zoph, Bojanowski, Ozyurt, Hedayatnia, Neyshabur, Inden, Stein, Ekmekci, Lin, Howald, Orinion, Diao, Dour, Stinson, Argueta, Ram'irez, Singh, Rathkopf, Meng, Baral, Wu, Callison-Burch, Waites, Voigt, Manning, Potts, Ramirez, Rivera, Siro, Raffel, Ashcraft, Garbacea, Sileo, Garrette, Hendrycks, Kilman, Roth, Freeman, Khashabi, Levy, Gonz'alez, Perszyk, Hernandez, Chen, Ippolito, Gilboa, Dohan, Drakard, Jurgens, Datta, Ganguli, Emelin, Kleyko, Yuret, Chen, Tam, Hupkes, Misra, Buzan, Mollo, Yang, Lee, Schrader,
  Shutova, Cubuk, Segal, Hagerman, Barnes, Donoway, Pavlick, Rodol{\`a}, Lam, Chu, Tang, Erdem, Chang, Chi, Dyer, Jerzak, Kim, Manyasi, Zheltonozhskii, Xia, Siar, Mart'inez-Plumed, Happ'e, Chollet, Rong, Mishra, Winata, de~Melo, Kruszewski, Parascandolo, Mariani, Wang, Jaimovitch-L'opez, Betz, Gur-Ari, Galijasevic, Kim, Rashkin, Hajishirzi, Mehta, Bogar, Shevlin, Schutze, Yakura, Zhang, Wong, Ng, Noble, Jumelet, Geissinger, Kernion, Hilton, Lee, Fisac, Simon, Koppel, Zheng, Zou, Koco'n, Thompson, Wingfield, Kaplan, Radom, Sohl-Dickstein, Phang, Wei, Yosinski, Novikova, Bosscher, Marsh, Kim, Taal, Engel, Alabi, Xu, Song, Tang, Waweru, Burden, Miller, Balis, Batchelder, Berant, Frohberg, Rozen, Hern{\'a}ndez-Orallo, Boudeman, Guerr, Jones, Tenenbaum, Rule, Chua, Kanclerz, Livescu, Krauth, Gopalakrishnan, Ignatyeva, Markert, Dhole, Gimpel, Omondi, Mathewson, Chiafullo, Shkaruta, Shridhar, McDonell, Richardson, Reynolds, Gao, Zhang, Dugan, Qin, Contreras-Ochando, philippe Morency, Moschella, Lam, Noble, Schmidt,
  He, Col'on, Metz, cSenel, Bosma, Sap, ter Hoeve, Farooqi, Faruqui, Mazeika, Baturan, Marelli, Maru, Quintana, Tolkiehn, Giulianelli, Lewis, Potthast, Leavitt, Hagen, Schubert, Baitemirova, Arnaud, McElrath, Yee, Cohen, Gu, Ivanitskiy, Starritt, Strube, Swkedrowski, Bevilacqua, Yasunaga, Kale, Cain, Xu, Suzgun, Walker, Tiwari, Bansal, Aminnaseri, Geva, Gheini, MukundVarma, Peng, Chi, Lee, Krakover, Cameron, Roberts, Doiron, Martinez, Nangia, Deckers, Muennighoff, Keskar, Iyer, Constant, Fiedel, Wen, Zhang, Agha, Elbaghdadi, Levy, Evans, Casares, Doshi, Fung, Liang, Vicol, Alipoormolabashi, Liao, Liang, Chang, Eckersley, Htut, Hwang, Milkowski, Patil, Pezeshkpour, Oli, Mei, Lyu, Chen, Banjade, Rudolph, Gabriel, Habacker, Risco, Milliere, Garg, Barnes, Saurous, Arakawa, Raymaekers, Frank, Sikand, Novak, Sitelew, Bras, Liu, Jacobs, Zhang, Salakhutdinov, Chi, Lee, Stovall, Teehan, Yang, Singh, Mohammad, Anand, Dillavou, Shleifer, Wiseman, Gruetter, Bowman, Schoenholz, Han, Kwatra, Rous, Ghazarian, Ghosh, Casey,
  Bischoff, Gehrmann, Schuster, Sadeghi, Hamdan, Zhou, Srivastava, Shi, Singh, Asaadi, Gu, Pachchigar, Toshniwal, Upadhyay, Debnath, Shakeri, Thormeyer, Melzi, Reddy, Makini, Lee, Torene, Hatwar, Dehaene, Divic, Ermon, Biderman, Lin, Prasad, Piantadosi, Shieber, Misherghi, Kiritchenko, Mishra, Linzen, Schuster, Li, Yu, Ali, Hashimoto, Wu, Desbordes, Rothschild, Phan, Wang, Nkinyili, Schick, Kornev, Tunduny, Gerstenberg, Chang, Neeraj, Khot, Shultz, Shaham, Misra, Demberg, Nyamai, Raunak, Ramasesh, Prabhu, Padmakumar, Srikumar, Fedus, Saunders, Zhang, Vossen, Ren, Tong, Zhao, Wu, Shen, Yaghoobzadeh, Lakretz, Song, Bahri, Choi, Yang, Hao, Chen, Belinkov, Hou, Hou, Bai, Seid, Zhao, Wang, Wang, Wang, and Wu}]{Srivastava2022BeyondTI}
Aarohi Srivastava, Abhinav Rastogi, Abhishek Rao, Abu Awal~Md Shoeb, Abubakar Abid, Adam Fisch, Adam~R. Brown, Adam Santoro, Aditya Gupta, Adri{\`a} Garriga-Alonso, Agnieszka Kluska, Aitor Lewkowycz, Akshat Agarwal, Alethea Power, Alex Ray, Alex Warstadt, Alexander~W. Kocurek, Ali Safaya, Ali Tazarv, and 431 others. 2022.
\newblock \href {https://api.semanticscholar.org/CorpusID:263625818} {Beyond the imitation game: Quantifying and extrapolating the capabilities of language models}.
\newblock \emph{ArXiv}, abs/2206.04615.

\bibitem[{Sun et~al.(2024)Sun, Xu, Yu, Chen, He, Zhao, and Liu}]{Sun2024ItDLL}
Wangtao Sun, Haotian Xu, Xuanqing Yu, Pei Chen, Shizhu He, Jun Zhao, and Kang Liu. 2024.
\newblock \href {https://api.semanticscholar.org/CorpusID:268357293} {Itd: Large language models can teach themselves induction through deduction}.
\newblock In \emph{Annual Meeting of the Association for Computational Linguistics}.

\bibitem[{Suzgun and Kalai(2024)}]{Suzgun2024MetaPromptingEL}
Mirac Suzgun and Adam~Tauman Kalai. 2024.
\newblock \href {https://api.semanticscholar.org/CorpusID:267094917} {Meta-prompting: Enhancing language models with task-agnostic scaffolding}.
\newblock \emph{ArXiv}, abs/2401.12954.

\bibitem[{Suzgun et~al.(2022)Suzgun, Scales, Scharli, Gehrmann, Tay, Chung, Chowdhery, Le, hsin Chi, Zhou, and Wei}]{Suzgun2022ChallengingBT}
Mirac Suzgun, Nathan Scales, Nathanael Scharli, Sebastian Gehrmann, Yi~Tay, Hyung~Won Chung, Aakanksha Chowdhery, Quoc~V. Le, Ed~Huai hsin Chi, Denny Zhou, and Jason Wei. 2022.
\newblock \href {https://api.semanticscholar.org/CorpusID:252917648} {Challenging big-bench tasks and whether chain-of-thought can solve them}.
\newblock In \emph{Annual Meeting of the Association for Computational Linguistics}.

\bibitem[{{Together AI}(2025)}]{together2025pricing}
{Together AI}. 2025.
\newblock \href {https://www.together.ai/pricing} {Together ai pricing}.
\newblock Accessed: May 2025.

\bibitem[{Wang et~al.(2023{\natexlab{a}})Wang, Xu, Lan, Hu, Lan, Lee, and Lim}]{wang-etal-2023-plan}
Lei Wang, Wanyu Xu, Yihuai Lan, Zhiqiang Hu, Yunshi Lan, Roy Ka-Wei Lee, and Ee-Peng Lim. 2023{\natexlab{a}}.
\newblock \href {https://doi.org/10.18653/v1/2023.acl-long.147} {Plan-and-solve prompting: Improving zero-shot chain-of-thought reasoning by large language models}.
\newblock In \emph{Proceedings of the 61st Annual Meeting of the Association for Computational Linguistics (Volume 1: Long Papers)}, pages 2609--2634, Toronto, Canada. Association for Computational Linguistics.

\bibitem[{Wang et~al.(2023{\natexlab{b}})Wang, Zelikman, Poesia, Pu, Haber, and Goodman}]{Wang2023HypothesisSI}
Ruocheng Wang, E.~Zelikman, Gabriel Poesia, Yewen Pu, Nick Haber, and Noah~D. Goodman. 2023{\natexlab{b}}.
\newblock \href {https://api.semanticscholar.org/CorpusID:261696510} {Hypothesis search: Inductive reasoning with language models}.
\newblock \emph{ArXiv}, abs/2309.05660.

\bibitem[{Wang et~al.(2022)Wang, Wei, Schuurmans, Le, Chi, and Zhou}]{Wang2022SelfConsistencyIC}
Xuezhi Wang, Jason Wei, Dale Schuurmans, Quoc Le, Ed~H. Chi, and Denny Zhou. 2022.
\newblock \href {https://api.semanticscholar.org/CorpusID:247595263} {Self-consistency improves chain of thought reasoning in language models}.
\newblock \emph{ArXiv}, abs/2203.11171.

\bibitem[{Wang et~al.(2024)Wang, Zhao, Wang, Huang, Fan, Zhang, Wang, Wang, and Liu}]{Wang2024StrategicCG}
Yu~Wang, Shiwan Zhao, Zhihu Wang, Heyuan Huang, Ming Fan, Yubo Zhang, Zhixing Wang, Haijun Wang, and Ting Liu. 2024.
\newblock \href {https://api.semanticscholar.org/CorpusID:272423721} {Strategic chain-of-thought: Guiding accurate reasoning in llms through strategy elicitation}.
\newblock \emph{ArXiv}, abs/2409.03271.

\bibitem[{Wei et~al.(2022)Wei, Wang, Schuurmans, Bosma, Chi, Xia, Le, and Zhou}]{Wei2022ChainOT}
Jason Wei, Xuezhi Wang, Dale Schuurmans, Maarten Bosma, Ed~H. Chi, F.~Xia, Quoc Le, and Denny Zhou. 2022.
\newblock \href {https://api.semanticscholar.org/CorpusID:246411621} {Chain of thought prompting elicits reasoning in large language models}.
\newblock \emph{ArXiv}, abs/2201.11903.

\bibitem[{White et~al.(2025)White, Dooley, Roberts, Pal, Feuer, Jain, Shwartz-Ziv, Jain, Saifullah, Dey, Shubh-Agrawal, Sandha, Naidu, Hegde, LeCun, Goldstein, Neiswanger, and Goldblum}]{livebench}
Colin White, Samuel Dooley, Manley Roberts, Arka Pal, Benjamin Feuer, Siddhartha Jain, Ravid Shwartz-Ziv, Neel Jain, Khalid Saifullah, Sreemanti Dey, Shubh-Agrawal, Sandeep~Singh Sandha, Siddartha~Venkat Naidu, Chinmay Hegde, Yann LeCun, Tom Goldstein, Willie Neiswanger, and Micah Goldblum. 2025.
\newblock Livebench: A challenging, contamination-free {LLM} benchmark.
\newblock In \emph{The Thirteenth International Conference on Learning Representations}.

\bibitem[{xAI(2025)}]{xAI_grok_3}
xAI. 2025.
\newblock \href {https://x.ai/news/grok-3} {Grok 3 beta — the age of reasoning agents}.
\newblock February 19, 2025.

\bibitem[{Yang et~al.(2022)Yang, Dong, Du, Cheng, Cambria, Liu, Gao, and Wei}]{Yang2022LanguageMA}
Zonglin Yang, Li~Dong, Xinya Du, Hao Cheng, E.~Cambria, Xiaodong Liu, Jianfeng Gao, and Furu Wei. 2022.
\newblock \href {https://api.semanticscholar.org/CorpusID:254926851} {Language models as inductive reasoners}.
\newblock \emph{ArXiv}, abs/2212.10923.

\bibitem[{Yao et~al.(2023)Yao, Yu, Zhao, Shafran, Griffiths, Cao, and Narasimhan}]{10.5555/3666122.3666639}
Shunyu Yao, Dian Yu, Jeffrey Zhao, Izhak Shafran, Thomas~L. Griffiths, Yuan Cao, and Karthik Narasimhan. 2023.
\newblock Tree of thoughts: deliberate problem solving with large language models.
\newblock In \emph{Proceedings of the 37th International Conference on Neural Information Processing Systems}, NIPS '23, Red Hook, NY, USA. Curran Associates Inc.

\bibitem[{Zheng et~al.(2023)Zheng, Mishra, Chen, Cheng, hsin Chi, Le, and Zhou}]{Zheng2023TakeAS}
Huaixiu~Steven Zheng, Swaroop Mishra, Xinyun Chen, Heng-Tze Cheng, Ed~Huai hsin Chi, Quoc~V. Le, and Denny Zhou. 2023.
\newblock \href {https://api.semanticscholar.org/CorpusID:263830368} {Take a step back: Evoking reasoning via abstraction in large language models}.
\newblock \emph{ArXiv}, abs/2310.06117.

\bibitem[{Zhu et~al.(2023)Zhu, Xue, Chen, Zhou, Tang, Schuurmans, and Dai}]{Zhu2023LargeLM}
Zhaocheng Zhu, Yuan Xue, Xinyun Chen, Denny Zhou, Jian Tang, Dale Schuurmans, and Hanjun Dai. 2023.
\newblock \href {https://api.semanticscholar.org/CorpusID:263834843} {Large language models can learn rules}.
\newblock \emph{ArXiv}, abs/2310.07064.

\end{thebibliography}
\clearpage

\appendix

\section{Experimental Configurations}
\label{sec:exp_config}

\begin{table*}[t!]
\centering
\small
\begin{tabular}{cccc}

\hline
\textbf{Model} & \textbf{\texttt{Model Version}} & \textbf{API Provider} & \textbf{cost} \\
\hline
Llama 3.3 8B~\cite{meta_2025_llama3_3} & \texttt{Llama-3.3-8B-Instruct} & Meta Llama API & - \\

Llama 3.3 70B~\cite{Dubey2024TheL3} & \texttt{Llama-3.3-70B-Instruct}& Meta Llama API & - \\

Llama 4 Scout~\cite{meta_2025_llama4} & \texttt{Llama-4-Scout-17B-16E-Instruct-FP8}& Meta Llama API & -  \\

Llama 4 Maverick~\cite{meta_2025_llama4} & \texttt{Llama-4-Maverick-17B-128E-Instruct-FP8}& Meta Llama API & - \\
\hline
Mistral Small 2~\cite{mistral2024small2} & \texttt{mistral-small-2409} & Mistral AI & 3\\

Mistral Small 3~\cite{mistral2025small3} & \texttt{mistral-small-2501} & Mistral AI & 3 \\

Mistral Small 3.1~\cite{mistral2025small3_1} & \texttt{mistral-small-2503} & Mistral AI & 3 \\

Mistral Medium 3~\cite{mistral_2025_medium_3} & \texttt{mistral-medium-2505} & Mistral AI & 25\\

Mistral Large 2~\cite{mistral2024Pixtral_Large} & \texttt{mistral-large-2411} & Mistral AI & 75\\
\hline
\shortstack{Gemini 2.0 Flash Lite}~\cite{google2025gemini2} & \texttt{gemini-2.0-flash-lite-001} & Google & 8 \\

Gemini 2.0 Flash~\cite{google2025gemini2} & \texttt{gemini-2.0-flash-001} & Google & 10 \\
\hline
GPT 4o mini~\cite{openai2024gpt4omini} & \texttt{gpt-4o-mini-2024-07-18} & OpenAI & 6\\

GPT 4o~\cite{openai2024gpt4o} & \texttt{gpt-4o-2024-11-20} & OpenAI & 100\\

GPT 4.1 nano~\cite{openai_2025_GPT_4_1} & \texttt{gpt-4.1-nano-2025-04-1} & OpenAI & 4\\

GPT 4.1 mini~\cite{openai_2025_GPT_4_1} & \texttt{gpt-4.1-mini-2025-04-14} & OpenAI & 16 \\

GPT 4.1~\cite{openai_2025_GPT_4_1} & \texttt{gpt-4.1-2025-04-14} & OpenAI & 80\\

o4 mini~\cite{openai_2025_o4_mini} & \texttt{o4-mini-2025-04-16} & OpenAI & 125 \\
\hline
Grok 3 Beta~\cite{xAI_grok_3} & \texttt{grok-3-beta} & xAI  & 200\\

Grok 3 Mini Beta~\cite{xAI_grok_3} & \texttt{grok-3-mini-beta} & xAI & 40\\
\hline
\end{tabular}
\caption{
Mapping of model names to detailed versions, API provider, and relative API cost (in prompt units; lower means cheaper). 
Note: At the time of writing, the \texttt{Meta Llama API} was in preview and provided free of charge.
}
\label{table:model_detail}
\end{table*}

\paragraph{Model List and Cost.}
Table~\ref{table:model_detail} details all model names, exact versions, API providers, and their relative API costs as used in our experiments. The cost column reflects the relative pricing or prompt quota at the time of writing. 

Table~\ref{table:model_detail} summarizes the estimated total cost for all experiments, which was approximately \$698, based on the providers' pricing during the experimental period.

\paragraph{Model Settings.}
For all experiments, we set the temperature to 0 and Top-P to 1, whenever these options were supported by the model.

\paragraph{Other Settings.}
To facilitate fair ablation studies on different numbers of rationales and draft plans, we designed our experiments such that all rationales and their corresponding draft plans were generated together in a single batch. Specifically, we prompted the model to produce $N=9$ rationales and their associated draft plans at once. For each ablation setting, we used only the first $k$ rationales and their corresponding draft plans for induction, ensuring that the initial $k$ inputs were always identical across settings. This setup guarantees that any variation in performance can be attributed solely to the number of rationales and draft plans, rather than differences in their content.

\section{Use of AI Assistants}

We used ChatGPT for grammar refinement and occasional assistance in code development. All language model outputs served solely as references; the final writing and code were thoroughly reviewed and authored by the authors.

\section{Prompt Templates and Examples}

\begin{figure*}[t]
  \centering
  \includegraphics[width=\linewidth]{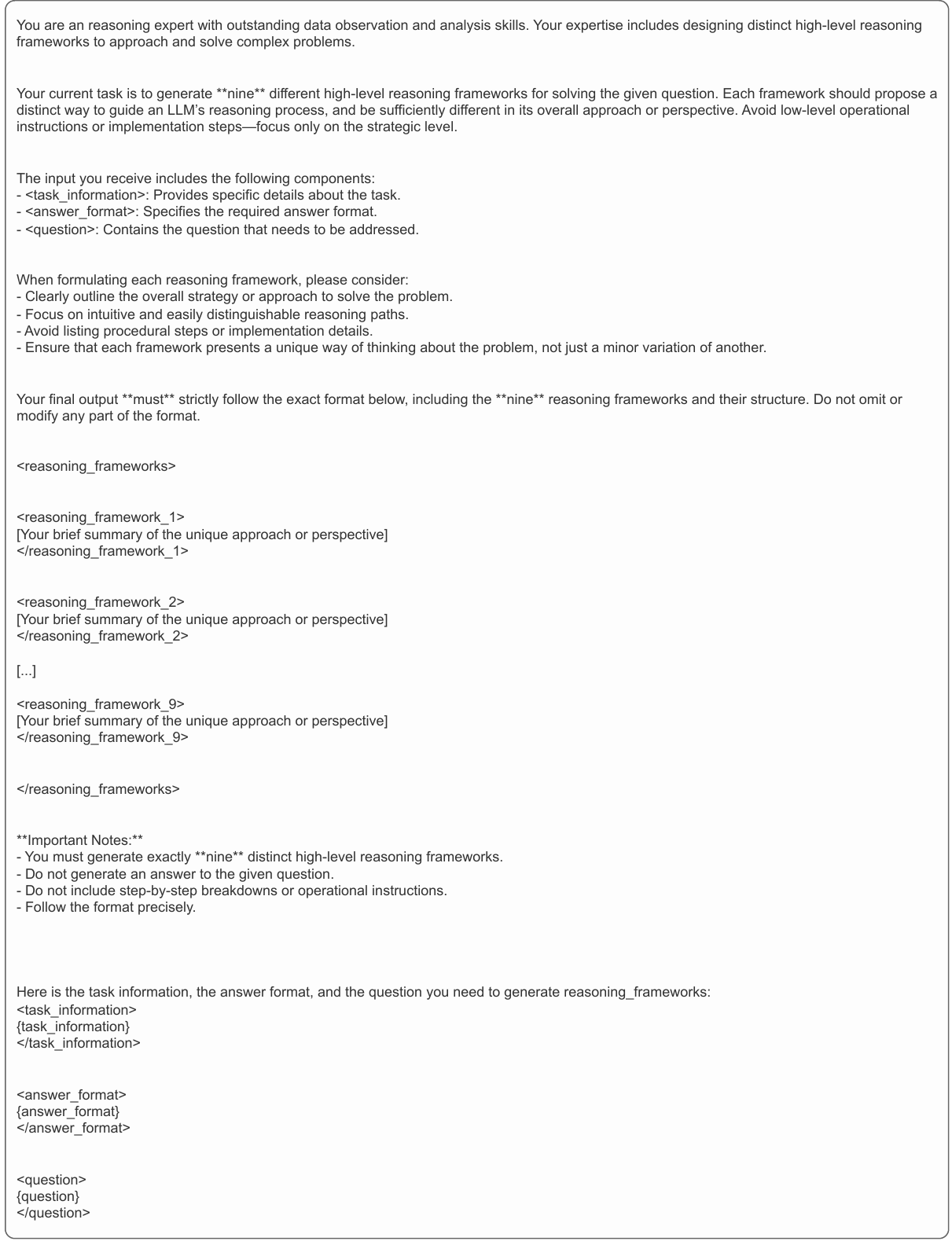}
    \caption{Prompt template used for Rationale Generation in the DIP framework.}
  \label{tab:DIP_Prompt_template_1}
\end{figure*}

\begin{figure*}[t]
  \centering
  \includegraphics[width=\linewidth]{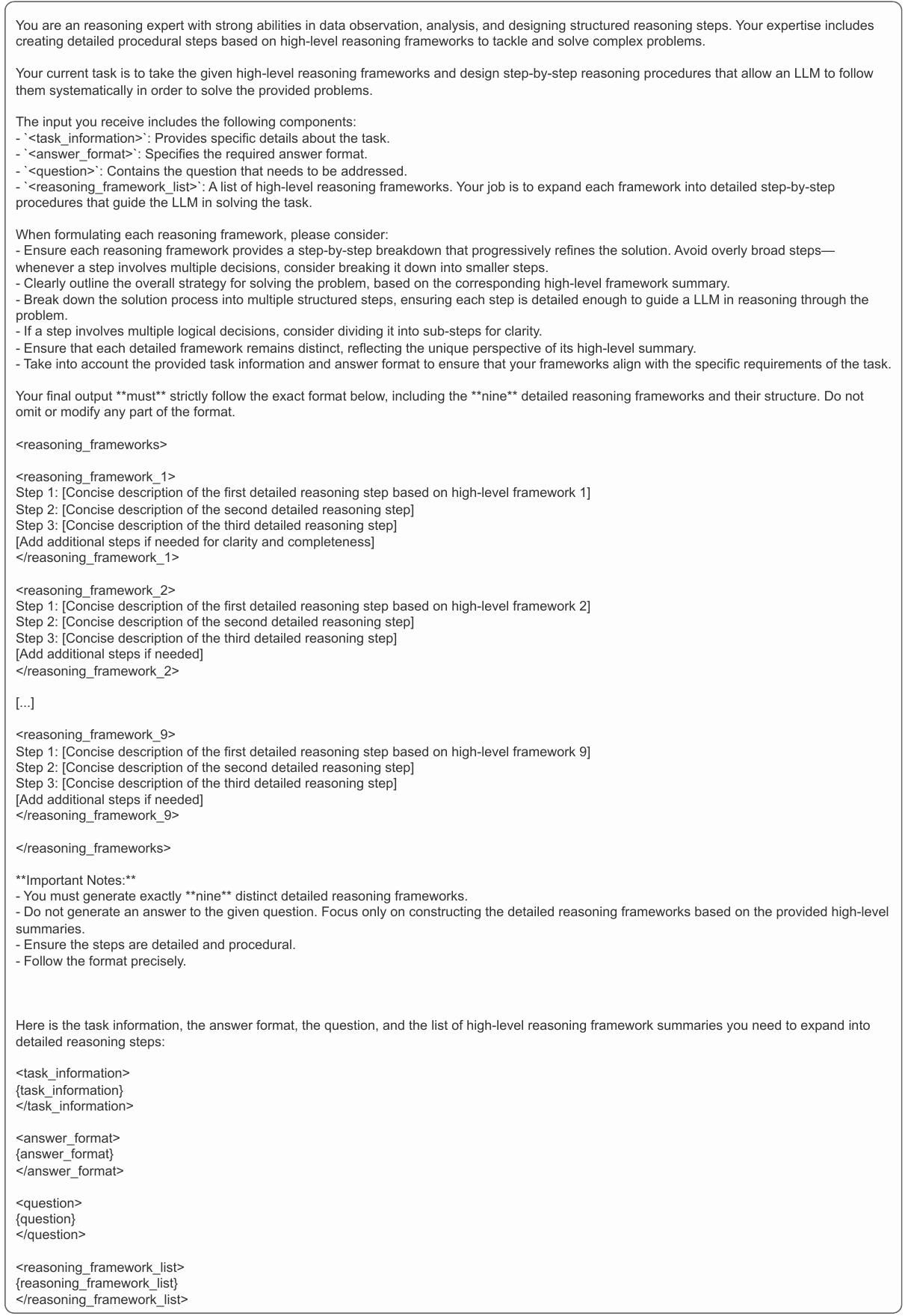}
    \caption{Prompt template used for Draft Plan Generation in the DIP framework.}
  \label{tab:DIP_Prompt_template_2}
\end{figure*}

\begin{figure*}[t]
  \centering
  \includegraphics[width=\linewidth]{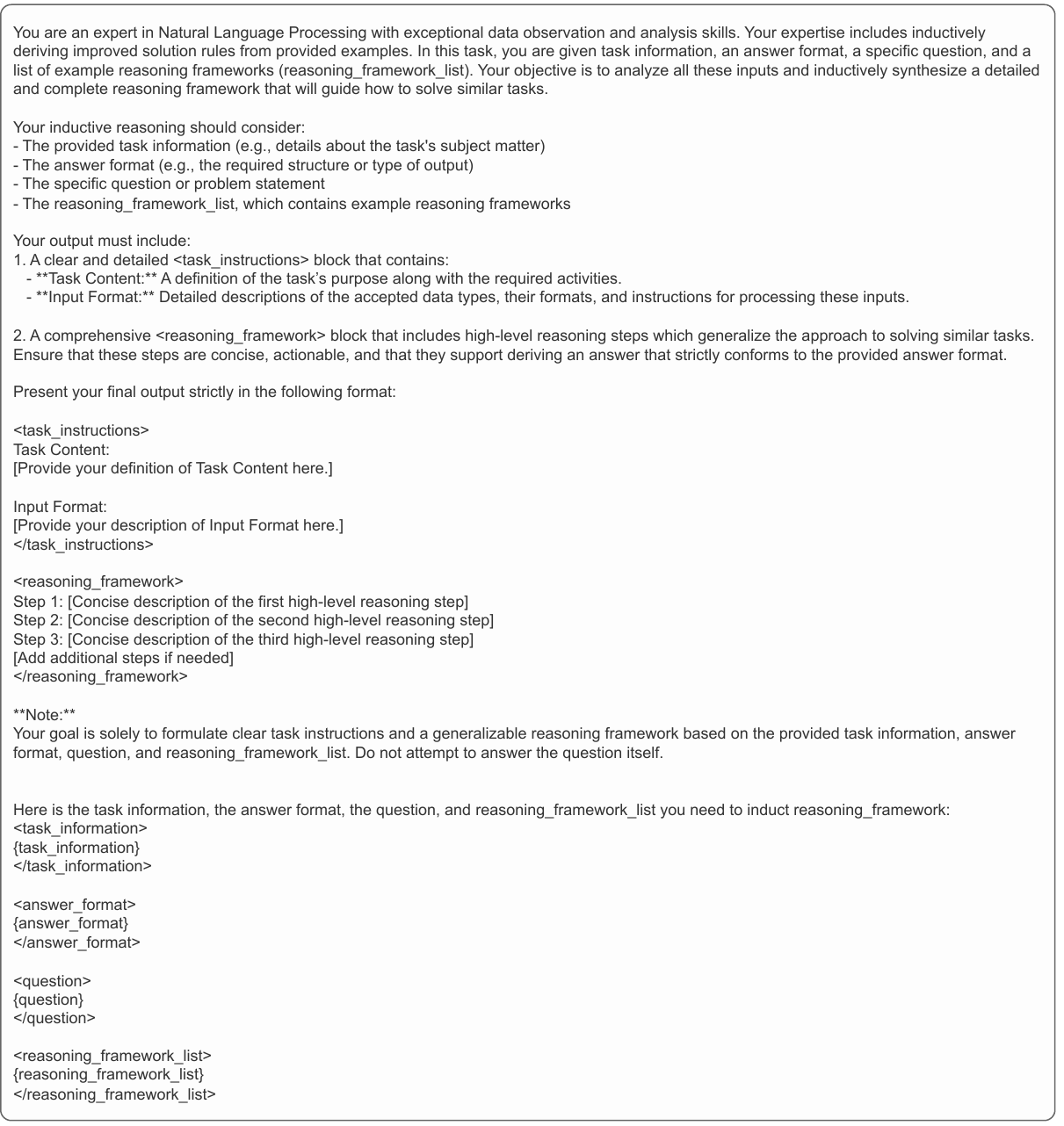}
    \caption{Prompt template used for Draft Plan Induction in the DIP framework.}
  \label{tab:DIP_Prompt_template_3}
\end{figure*}

\begin{figure*}[t]
  \centering
  \includegraphics[width=\linewidth]{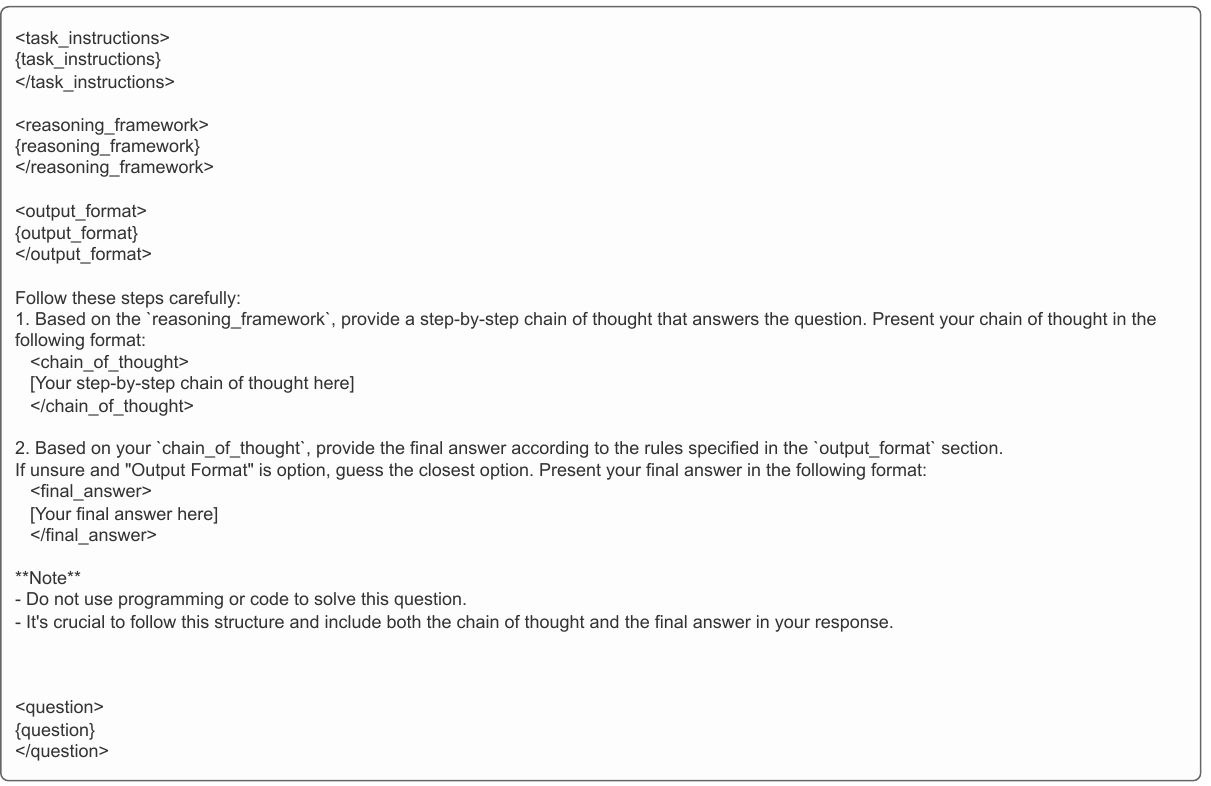}
    \caption{Prompt template used for Answer Generation in the DIP framework.}
  \label{tab:DIP_Prompt_template_4}
\end{figure*}

Figures~\ref{tab:DIP_Prompt_template_1} to \ref{tab:DIP_Prompt_template_4} present the exact prompt templates used at each stage of our DIP framework, covering Rationale Generation, Draft Plan Construction, Draft Plan Induction, and Answer Generation. Each template is provided in full to ensure transparency and reproducibility.

Figures~\ref{sec:full-case-Reasoning-Approaches} to \ref{tab:RCOTTranslationError} contain the complete case study examples referenced in the main paper. These examples include all intermediate outputs from the ablation variants and baseline methods, as evaluated using the \texttt{Llama 4 Scout} model. Only the complete DIP method produces the correct answer for these challenging cases, while all other ablations and baselines fail to do so.

\begin{figure*}[ht]
  \small
  \centering
  \begin{tabularx}{\textwidth}{>{\raggedright\arraybackslash}p{2.6cm} X}
    \toprule
    \textbf{Method} & \textbf{Rationales or Draft Plan} \\
    \midrule

    DIP N=5 &
    {\scriptsize
    \textbf{\textcolor{blue}{Rationales}}\par
    \begin{enumerate}[leftmargin=*,nosep,label=(\arabic*)]
      \item Analyze the translation by comparing the source and target sentences to identify changes in named entities, numerical values, modifiers, negation, facts, or dropped content.
      \item Focus on identifying changes in the family of the snake in the source and translation sentences to determine the type of error.
      \item Examine name changes such as ``Mexikanische Königsnatter'' $\rightarrow$ ``Mexican King's Snake'' and ``San-Luis-Potosi-Königsnatter'' $\rightarrow$ ``San Luis Potosi King snake''.
      \item Investigate whether the adjective/modifier for the family term shifts from ``Nattern'' to ``Lantern family''.
      \item Check for any negation or antonyms introduced in the translation that could change the sentence meaning.
    \end{enumerate}

    {\color{gray}\rule{\linewidth}{0.4pt}}\par
    \textbf{\textcolor{blue}{Draft Plan}}\par
    \begin{enumerate}[leftmargin=*,nosep,label=\textbf{Plan \arabic*:}]
      \item Step 1: Compare the source sentence ``Die Mexikanische Königsnatter …'' with the translation ``The Mexican King's Snake …'' to identify changes.\newline
            Step 2: Analyze the translation for changes in named entities (snake and family names).\newline
            Step 3: Check for changes in numerical values, modifiers, negation, facts, or dropped content.\newline
            Step 4: Determine the error type based on the identified changes.
      \item Step 1: Identify the family of the snake in the source sentence (``der Nattern'').\newline
            Step 2: Identify the family of the snake in the translation (``the Lantern family'').\newline
            Step 3: Compare the two families to determine if there is a change.\newline
            Step 4: If a change exists, classify it as an error in named entities or modifiers/adjectives.
      \item Step 1: Examine the translation of ``Mexikanische Königsnatter'' $\rightarrow$ ``Mexican King's Snake'' for named-entity changes.\newline
            Step 2: Examine ``San-Luis-Potosi-Königsnatter'' $\rightarrow$ ``San Luis Potosi King snake'' for named-entity changes.\newline
            Step 3: Decide if the changes are significant enough to be classified as a named-entity error.
      \item Step 1: Identify the modifier/adjective for the family term in the source (``Nattern'').\newline
            Step 2: Identify the modifier/adjective for the family term in the translation (``Lantern family'').\newline
            Step 3: Compare the two modifiers/adjectives.\newline
            Step 4: Classify the change as a modifier/adjective error.
      \item Step 1: Check the source sentence for any negations or comparatives.\newline
            Step 2: Check the translation for any negations or comparatives.\newline
            Step 3: Compare both sentences to determine if there is a change in negation or antonyms.
    \end{enumerate}}\\[8pt]

    \midrule
    DIP-R N=5 &
    {\scriptsize
    \textbf{\textcolor{blue}{Draft Plan}}\par
    \begin{enumerate}[leftmargin=*,nosep,label=\textbf{Plan \arabic*:}]
      \item Step 1: Identify the key elements in the source sentence and the translation.\newline
            Step 2: Compare the family names (``Familie der Nattern'' vs.\ ``Lantern family'').\newline
            Step 3: Determine the type of error based on this comparison.
      \item Step 1: Analyze the source sentence for specific details.\newline
            Step 2: Examine the translation for changes in named entities, numerical values, modifiers, negation, or dropped content.\newline
            Step 3: Classify the error based on the identified changes.
      \item Step 1: Check for changes in named entities between the source and translation.\newline
            Step 2: Verify changes in numerical values, modifiers, or negation.\newline
            Step 3: Identify if any content has been dropped or if there is a factual error.
      \item Step 1: Compare the source sentence and translation for changes in names or locations.\newline
            Step 2: Examine whether there are changes in numerical values or units.\newline
            Step 3: Determine if the error involves modifiers, negation, or factual errors.
      \item Step 1: Identify the error by comparing the family names in both sentences.\newline
            Step 2: Classify the type of error (e.g., named entities, numerical values, modifiers).\newline
            Step 3: Confirm the error type based on the comparison.
    \end{enumerate}}\\
    \bottomrule
  \end{tabularx}
    \caption{
    BBH Salient Translation Error Detection example: All rationales and plans generated by DIP ($N=5$) and DIP-R ($N=5$), where DIP-R omits the Rationale Generation step.
    }
    \label{sec:full-case-Reasoning-Approaches}
\end{figure*}

\begin{figure*}[ht]
  \small
  \centering  
  \begin{tabularx}{\textwidth}{p{1.9cm} X}
    \toprule
    \textbf{Method} & \textbf{Inducted Draft Plan} \\
    \midrule

    DIP (N=5) &
    \begin{minipage}[t]{\linewidth}\scriptsize
      Step 1: Compare the source sentence ``Die~Mexikanische~Königsnatter oder San-Luis-Potosi-Königsnatter ist eine Schlange aus der Familie der Nattern'' with the translation ``The Mexican King's Snake or San Luis Potosi King snake is a snake of the Lantern family'' to identify any changes.\\[2pt]
      Step 2: Analyze the translation for changes in named entities, such as the names of the snake and its family.\\[2pt]
      Step 3: Check for changes in numerical values, modifiers, negation, facts, or dropped content.\\[2pt]
      Step 4: Determine the type of error based on the identified changes.\\[4pt]
      More specifically: \\
      \textendash{} Identify the family of the snake in the source sentence (``der Nattern'') and in the translation (``the Lantern family'').\\
      \textendash{} Compare the two families to determine if there is a change.\\
      \textendash{} Examine the translation of ``Mexikanische Königsnatter'' to ``Mexican King's Snake'' and ``San-Luis-Potosi-Königsnatter'' to ``San Luis Potosi King snake'' for any changes in named entities.\\
      \textendash{} Check if the changes are significant enough to be classified as an error in named entities or modifiers/adjectives.\\
      \textendash{} Verify if there are any changes in negation or antonyms.\\
      \textendash{} Verify if there are any changes in numerical values or facts.\\
      \textendash{} Verify if any significant clause is dropped in the translation.\\[4pt]
      Based on these steps, classify the error into one of the provided options: (A) Modifiers or Adjectives, (B) Numerical Values, (C) Negation or Antonyms, (D) Named Entities, (E) Dropped Content, (F) Facts.
    \end{minipage}
    \\ \addlinespace[6pt]

    \midrule
    DIP-R (N=5) &
    \begin{minipage}[t]{\linewidth}\scriptsize
      Step 1: Identify the key elements in the source sentence and the translation.\\[2pt]
      Step 2: Compare the key elements such as family names, entities, numerical values, modifiers, and content to identify any changes.\\[2pt]
      Step 3: Determine the type of error based on the comparison, considering the categories: Named Entities, Numerical Values, Modifiers or Adjectives, Negation or Antonyms, Facts, and Dropped Content.\\[2pt]
      Step 4: Classify the error into one of the provided options (A, B, C, D, E, F) based on the identified change.\\[2pt]
      Step 5: Confirm the error type based on the comparison and classification.
    \end{minipage}
    \\ \addlinespace[6pt]

    \midrule
    DIP (N=1) &
    \begin{minipage}[t]{\linewidth}\scriptsize
      Step 1: Compare the source sentence with the translation to identify any changes in terms of named entities, numerical values, modifiers or adjectives, negation or antonyms, facts, and dropped content.\\[2pt]
      Step 2: Analyze the translation for changes in named entities, such as the names of the snake and its family, and check if they match the source sentence.\\[2pt]
      Step 3: Check for changes in numerical values, modifiers, negation, facts, or dropped content by carefully examining the source and translated sentences.\\[2pt]
      Step 4: Determine the type of error based on the identified changes by matching them with the given error types: Named Entities, Numerical Values, Modifiers or Adjectives, Negation or Antonyms, Facts, and Dropped Content.\\[2pt]
      Step 5: Select the correct option from the provided list that corresponds to the identified error type.
    \end{minipage}
    \\
    \bottomrule    
  \end{tabularx}
    \caption{
    BBH Salient Translation Error Detection example: Final induced draft plans for each method.
    }
    \label{sec:full-case}
\end{figure*}

\begin{figure*}[ht!]
  \centering
  {\tiny  
  \begin{tabular}{p{0.97\linewidth}}
  \hline
    \textcolor{blue}{DIP Prompt} \\
    \texttt{<task\_instructions>} \\
    Task Content: \\
    The task requires detecting the type of error in an English translation of a German source sentence. The error types include Named Entities, Numerical Values, Modifiers or Adjectives, Negation or Antonyms, Facts, and Dropped Content. \\
    Input Format: \\
    - The source sentence in German \\
    - The translated sentence in English \\
    - A list of possible error types \\
    - A set of answer options corresponding to the error types \\
    The task involves comparing the source sentence and the translation to identify any changes and determining the type of error based on the identified changes. \\
    \texttt{</task\_instructions>} \\[1ex]
    \textcolor{blue}{Reasoning Framework} \\
    \texttt{<reasoning\_framework>} \\
    Step 1: Compare the source sentence "Die Mexikanische Königsnatter oder San-Luis-Potosi-Königsnatter ist eine Schlange aus der Familie der Nattern" with the translation "The Mexican King's Snake or San Luis Potosi King snake is a snake of the Lantern family" to identify any changes. \\
    Step 2: Analyze the translation for changes in named entities, such as the names of the snake and its family. \\
    Step 3: Check for changes in numerical values, modifiers, negation, facts, or dropped content. \\
    Step 4: Determine the type of error based on the identified changes. \\
    More specifically: \\
    - Identify the family of the snake in the source sentence ("der Nattern") and in the translation ("the Lantern family"). \\
    - Compare the two families to determine if there is a change. \\
    - Examine the translation of "Mexikanische Königsnatter" to "Mexican King's Snake" and "San-Luis-Potosi-Königsnatter" to "San Luis Potosi King snake" for any changes in named entities. \\
    - Check if the changes are significant enough to be classified as an error in named entities or modifiers/adjectives. \\
    - Verify if there are any changes in negation or antonyms. \\
    - Verify if there are any changes in numerical values or facts. \\
    - Verify if any significant clause is dropped in the translation. \\
    Based on these steps, classify the error into one of the provided options: \\
    (A) Modifiers or Adjectives  (B) Numerical Values (C) Negation or Antonyms  (D) Named Entities  (E) Dropped Content  (F) Facts \\
    \texttt{</reasoning\_framework>} \\[1ex]
    \textcolor{blue}{Prompt Template} \\
    \texttt{<output\_format>} \\
    The output is a single letter corresponding to the correct option from the list (e.g., A, B, C, etc.). \\
    \texttt{</output\_format>} \\

    Follow these steps carefully: \\
    1. Based on the \texttt{reasoning\_framework}, provide a step-by-step chain of thought that answers the question. Present your chain of thought in the following format: \\
    \quad \texttt{<chain\_of\_thought>} \\
    \quad [Your step-by-step chain of thought here] \\
    \quad \texttt{</chain\_of\_thought>} \\

    2. Based on your \texttt{chain\_of\_thought}, provide the final answer according to the rules specified in the \texttt{output\_format} section. \\
    If unsure and "Output Format" is option, guess the closest option. Present your final answer in the following format: \\
    \quad \texttt{<final\_answer>} \\
    \quad [Your final answer here] \\
    \quad \texttt{</final\_answer>} \\

    \textbf{Note}: \\
    - Do not use programming or code to solve this question. \\
    - It's crucial to follow this structure and include both the chain of thought and the final answer in your response. \\
    \textcolor{blue}{Question}\\ 
    \texttt{<question>} \\
    The following translations from German to English contain a particular error. That error will be one of the following types: Named Entities: An entity (names, places, locations, etc.) is changed to a different entity. Numerical Values: Numerical values (ordinals or cardinals), dates, and/or units are changed. Modifiers or Adjectives: The modifiers and adjectives pertaining to a noun are changed. Negation or Antonyms: Introduce or remove a negation or change comparatives to their antonyms. Facts: Trivial factual errors not pertaining to the above classes are introduced in the translations. Dropped Content: A significant clause in the translation is removed. Please identify that error.  Source: Die Mexikanische Königsnatter oder San-Luis-Potosi-Königsnatter ist eine Schlange aus der Familie der Nattern. \\
    Translation: The Mexican King's Snake or San Luis Potosi King snake is a snake of the Lantern family. \\
    The translation contains an error pertaining to \\
    Options: \\
    (A) Modifiers or Adjectives  (B) Numerical Values  (C) Negation or Antonyms  (D) Named Entities  (E) Dropped Content  (F) Facts \\
    \texttt{</question>} \\[1ex]
    \textcolor{blue}{Model Response}\\  
    \texttt{<chain\_of\_thought>} \\
    1. \textbf{Identify the source sentence and the translation:} \\
    \quad - Source sentence: "Die Mexikanische Königsnatter oder San-Luis-Potosi-Königsnatter ist eine Schlange aus der Familie der Nattern." \\
    \quad - Translation: "The Mexican King's Snake or San Luis Potosi King snake is a snake of the Lantern family." \\
    2. \textbf{Compare named entities:} \\
    \quad - The source sentence mentions "Mexikanische Königsnatter" and "San-Luis-Potosi-Königsnatter" as the names of the snakes. \\
    \quad - The translation mentions "Mexican King's Snake" and "San Luis Potosi King snake," which are similar but not exactly the same due to capitalization and spacing differences, but essentially convey the same information. \\
    3. \textbf{Check for changes in family names:} \\
    \quad - The source sentence states the snake is from "der Familie der Nattern." \\
    \quad - The translation states the snake is from "the Lantern family." \\
    \quad - The family name has changed from "Nattern" to "Lantern family." \\
    4. \textbf{Determine the type of error:} \\
    \quad - The change from "der Nattern" (which translates to "the colubrids" or a family of snakes) to "the Lantern family" seems to be a change in a named entity or a factual piece of information about the snake's family. \\
    5. \textbf{Analyze each option:} \\
    \quad - \textbf{Named Entities (D):} The family name "Nattern" (colubrids) has been changed to "Lantern family," which could be considered a change in a named entity. \\
    \quad - \textbf{Modifiers or Adjectives (A):} There are no obvious changes in modifiers or adjectives that pertain to a noun in a way that would classify as an error here. \\
    \quad - \textbf{Numerical Values (B):} There are no numerical values mentioned in the source or translation. \\
    \quad - \textbf{Negation or Antonyms (C):} There is no introduction or removal of a negation. \\
    \quad - \textbf{Facts (F):} The change in the family name could also be considered a factual error. \\
    \quad - \textbf{Dropped Content (E):} No significant clause appears to be dropped. \\
    6. \textbf{Conclusion:} \\
    \quad - The most direct error is the change in the family name from "Nattern" to "Lantern family," which can be classified as a change in a named entity or a factual error. However, given that "Lantern family" is not a recognized family of snakes and "Nattern" refers to a specific family (colubrids), this seems to be a clear case of changing a named entity or introducing a factual error. \\
    7. \textbf{Final Classification:} \\
    \quad - Between \textbf{Named Entities (D)} and \textbf{Facts (F)}, the error seems to lean more towards \textbf{Named Entities (D)} because it involves changing the name of a specific family of snakes. \\
    \texttt{</chain\_of\_thought>} \\

    \texttt{<final\_answer>} \\
    \textcolor{red}{D} \\
    \texttt{</final\_answer>} \\
    \hline
  \end{tabular}
  }
\caption{Example of the DIP prompt (N=5) and reasoning for translation error detection in the BBH-Induct dataset. The gold (correct) answer for this example is (D) Named Entities.}
\label{tab:DIPTranslationErrorDetection}
\end{figure*}

\begin{table*}[ht]
  \small
  \centering
  \begin{tabular}{l|ccccc|ccccc|}
    \toprule
    \textbf{Model} & \multicolumn{5}{c}{\textbf{BBH}} & \multicolumn{5}{c}{\textbf{LiveBench Reasoning}} \\
     & \textbf{N=1} & \textbf{N=3} & \textbf{N=5} & \textbf{N=7} & \textbf{N=9} & \textbf{N=1} & \textbf{N=3} & \textbf{N=5} & \textbf{N=7} & \textbf{N=9} \\
    \midrule
    Llama 4 Scout & 83.65 & 84.12 & \textbf{84.46} & 82.38 & 82.20 & 43.00 & 44.00 & 43.00 & \textbf{46.00} & 44.00 \\
    Llama 4 Maverick & 86.96 & 86.20 & 86.20 & \textbf{87.77} & 87.48 & 55.00 & 56.50 & 56.50 & 56.00 & \textbf{60.50} \\
    \midrule
    Mistral Small 3 & 80.06 & 78.49 & 81.04 & \textbf{81.97} & 80.81 & 34.00 & 31.50 & 36.50 & \textbf{40.50} & 30.50 \\
    Mistral Medium 3 & 86.78 & 87.88 & 87.01 & 88.17 & \textbf{88.81} & 53.50 & 53.50 & 54.50 & \textbf{59.00} & 55.00 \\
    \midrule
    Gemini 2.0 Flash Lite & 81.04 & 83.71 & 83.54 & \textbf{84.17} & 83.42 & 43.50 & 44.50 & \textbf{48.00} & 45.50 & 43.50 \\
    Gemini 2.0 Flash & 85.10 & 83.65 & \textbf{85.22} & 82.55 & 82.55 & 50.00 & 54.00 & \textbf{58.50} & 57.50 & 54.50 \\
    \midrule
    GPT 4.1 Mini & 89.68 & 89.91 & 89.86 & \textbf{90.09} & 88.81 & 64.50 & 67.50 & 65.00 & \textbf{68.00} & 65.50 \\
    O4 Mini & 90.49 & 90.72 & \textbf{91.59} & 90.03 & 90.61 & 86.50 & 89.00 & \textbf{91.00} & 87.50 & 88.00 \\
    \midrule
    Grok 3 & 90.61 & 90.78 & 90.61 & 91.13 & \textbf{91.48} & 70.00 & 69.00 & \textbf{73.00} & 71.00 & 71.50 \\
    Grok 3 Mini (Low) & 89.39 & \textbf{91.83} & 90.26 & 91.13 & 91.30 & 82.00 & 82.50 & 84.50 & 83.50 & \textbf{85.50} \\
    \midrule
    Win N=1 Rate & - & 7/10 & 8/10 & 7/10 & 7/10 & - & 7/10 & 9/10 & 10/10 & 8/10 \\
    \bottomrule
  \end{tabular}
  \caption{Performance (\%) of different models on \textsc{BBH} and LiveBench Reasoning, as a function of the number of rationales $N$. For each $N>1$, the last row reports the number of models that outperform $N=1$.}
  \label{tab:nshot_performance_full}
\end{table*}

\begin{figure*}[ht!]
  \centering
  {\tiny  
  \begin{tabular}{p{0.97\linewidth}}
  \hline
    \textcolor{blue}{DIP Prompt} \\
    \texttt{<task\_instructions>} \\
    Task Content: \\
    The task requires detecting the type of error in an English translation of a German source sentence. The error types to be considered are: Named Entities, Numerical Values, Modifiers or Adjectives, Negation or Antonyms, Facts, and Dropped Content. \\
    Input Format: \\
    - The source sentence in German. \\
    - The translated sentence in English. \\
    - A list of possible error types. \\
    The task involves comparing the source sentence and the translation to identify the error type. The output is a single letter corresponding to the correct option from the list. \\
    \texttt{</task\_instructions>} \\[1ex]
    \\
    \textcolor{blue}{Reasoning Framework} \\
    \texttt{<reasoning\_framework>} \\
    Step 1: Identify the key elements in the source sentence and the translation. \\
    Step 2: Compare the key elements such as family names, entities, numerical values, modifiers, and content to identify any changes. \\
    Step 3: Determine the type of error based on the comparison, considering the categories: Named Entities, Numerical Values, Modifiers or Adjectives, Negation or Antonyms, Facts, and Dropped Content. \\
    Step 4: Classify the error into one of the provided options (A, B, C, D, E, F) based on the identified change. \\
    Step 5: Confirm the error type based on the comparison and classification. \\
    \texttt{</reasoning\_framework>} \\[1ex]
    \\
    \textcolor{blue}{Prompt Template} \\
    \texttt{<output\_format>} \\
    The output is a single letter corresponding to the correct option from the list (e.g., A, B, C, etc.). \\
    \texttt{</output\_format>} \\

    Follow these steps carefully: \\
    1. Based on the \texttt{reasoning\_framework}, provide a step-by-step chain of thought that answers the question. \\
    \quad \texttt{<chain\_of\_thought>} \\
    \quad [Your step-by-step chain of thought here] \\
    \quad \texttt{</chain\_of\_thought>} \\

    2. Based on your \texttt{chain\_of\_thought}, provide the final answer according to the rules specified in the \texttt{output\_format} section. \\
    If unsure and "Output Format" is option, guess the closest option. Present your final answer in the following format: \\
    \quad \texttt{<final\_answer>} \\
    \quad [Your final answer here] \\
    \quad \texttt{</final\_answer>} \\

    \textbf{Note} Do not use programming or code to solve this question. \\
    \\
    \textcolor{blue}{Question}\\ 
    \texttt{<question>} \\
    The following translations from German to English contain a particular error. That error will be one of the following types: Named Entities: An entity (names, places, locations, etc.) is changed to a different entity. Numerical Values: Numerical values (ordinals or cardinals), dates, and/or units are changed. Modifiers or Adjectives: The modifiers and adjectives pertaining to a noun are changed. Negation or Antonyms: Introduce or remove a negation or change comparatives to their antonyms. Facts: Trivial factual errors not pertaining to the above classes are introduced in the translations. Dropped Content: A significant clause in the translation is removed. Please identify that error.  Source: Die Mexikanische Königsnatter oder San-Luis-Potosi-Königsnatter ist eine Schlange aus der Familie der Nattern. \\
    Translation: The Mexican King's Snake or San Luis Potosi King snake is a snake of the Lantern family. \\
    The translation contains an error pertaining to \\
    Options: \\
    (A) Modifiers or Adjectives \\
    (B) Numerical Values \\
    (C) Negation or Antonyms \\
    (D) Named Entities \\
    (E) Dropped Content \\
    (F) Facts \\
    \texttt{</question>} \\[1ex]
    \\
    \textcolor{blue}{Model Response}\\  
    \texttt{<chain\_of\_thought>} \\
    To identify the error type in the translation, let's compare the source sentence and the translation step by step. \\
    1. \textbf{Source Sentence}: Die Mexikanische Königsnatter oder San-Luis-Potosi-Königsnatter ist eine Schlange aus der Familie der Nattern. \\
    2. \textbf{Translation}: The Mexican King's Snake or San Luis Potosi King snake is a snake of the Lantern family. \\
    \textbf{Step 1: Identify Key Elements} \\
    - Source sentence mentions "Mexikanische Königsnatter" and "San-Luis-Potosi-Königsnatter" as names. \\
    - Translation mentions "Mexican King's Snake" and "San Luis Potosi King snake" as names. \\
    - Source sentence states the family as "der Nattern." \\
    - Translation states the family as "the Lantern family." \\
    \textbf{Step 2: Compare Key Elements} \\
    - The names "Mexikanische Königsnatter" and "San-Luis-Potosi-Königsnatter" are translated to "Mexican King's Snake" and "San Luis Potosi King snake," respectively. This seems to be a change in the naming but let's check if it's an entity change or just a translation. \\
    - The family in the source sentence is "der Nattern" (which translates to "the colubrids" or a general term for snakes, specifically non-venomous snakes), and in the translation, it's "the Lantern family." \\
    \textbf{Step 3: Determine the Type of Error} \\
    - \textbf{Named Entities}: The names of the snakes are translated but not significantly altered in terms of entity type (still referring to the same snakes). However, "Lantern family" seems to be a different classification than "der Nattern." \\
    - \textbf{Numerical Values}: There's no numerical value mentioned in either sentence. \\
    - \textbf{Modifiers or Adjectives}: The modifiers are translated (Mexikanische to Mexican, Königsnatter to King's Snake), which seems accurate. \\
    - \textbf{Negation or Antonyms}: There's no introduction or removal of negation. \\
    - \textbf{Facts}: The mention of "Lantern family" instead of the actual family "Nattern" (or colubrids) seems to be a factual error regarding the classification of the snakes. \\
    - \textbf{Dropped Content}: No significant clause seems to be removed. \\
    \textbf{Step 4: Classify the Error} \\
    Given that "der Nattern" refers to a specific family of snakes (colubrids) and "Lantern family" is not a recognized family of snakes in this context (it seems to be a mistranslation or misclassification), this appears to be a factual error regarding the classification. \\
    \textbf{Step 5: Confirm the Error Type} \\
    The error is best classified as a \textbf{factual error} because it misrepresents the family of the snakes. This aligns with option \textbf{(F) Facts}. \\
    \texttt{</chain\_of\_thought>} \\

    \texttt{<final\_answer>} \\
    \textcolor{red}{F} \\
    \texttt{</final\_answer>} \\
    \hline
  \end{tabular}
  }
\caption{Example of the DIP-R prompt and reasoning for translation error detection in the BBH-Induct dataset, using a step-by-step framework. The gold (correct) answer for this example is (D) Named Entities.}
\label{tab:DIPRTranslationErrorDetection}
\end{figure*}

\begin{figure*}[ht!]
  \centering
  {\tiny  
  \begin{tabular}{p{0.97\linewidth}}
  \hline
    \textcolor{blue}{DIP Prompt (N=1)} \\
    \texttt{<task\_instructions>} \\
    Task Content: \\
    The task requires detecting the type of error in an English translation of a German source sentence. The error types to be considered are: Named Entities, Numerical Values, Modifiers or Adjectives, Negation or Antonyms, Facts, and Dropped Content. \\
    Input Format: \\
    - The source sentence in German. \\
    - The translated sentence in English. \\
    - A list of possible error types. \\
    The task involves comparing the source sentence with the translation to identify any changes and determining the type of error based on the identified changes. \\
    \texttt{</task\_instructions>} \\[1ex]
    \\
    \textcolor{blue}{Reasoning Framework} \\
    \texttt{<reasoning\_framework>} \\
    Step 1: Compare the source sentence with the translation to identify any changes in terms of named entities, numerical values, modifiers or adjectives, negation or antonyms, facts, and dropped content. \\
    Step 2: Analyze the translation for changes in named entities, such as the names of the snake and its family, and check if they match the source sentence. \\
    Step 3: Check for changes in numerical values, modifiers, negation, facts, or dropped content by carefully examining the source and translated sentences. \\
    Step 4: Determine the type of error based on the identified changes by matching them with the given error types: Named Entities, Numerical Values, Modifiers or Adjectives, Negation or Antonyms, Facts, and Dropped Content. \\
    Step 5: Select the correct option from the provided list that corresponds to the identified error type. \\
    \texttt{</reasoning\_framework>} \\[1ex]
    \\
    \textcolor{blue}{Prompt Template} \\
    \texttt{<output\_format>} \\
    The output is a single letter corresponding to the correct option from the list (e.g., A, B, C, etc.). \\
    \texttt{</output\_format>} \\

    Follow these steps carefully: \\
    1. Based on the \texttt{reasoning\_framework}, provide a step-by-step chain of thought that answers the question. Present your chain of thought in the following format: \\
    \quad \texttt{<chain\_of\_thought>} \\
    \quad [Your step-by-step chain of thought here] \\
    \quad \texttt{</chain\_of\_thought>} \\

    2. Based on your \texttt{chain\_of\_thought}, provide the final answer according to the rules specified in the \texttt{output\_format} section. \\
    If unsure and "Output Format" is option, guess the closest option. Present your final answer in the following format: \\
    \quad \texttt{<final\_answer>} \\
    \quad [Your final answer here] \\
    \quad \texttt{</final\_answer>} \\

    \textbf{Note} \\
    - Do not use programming or code to solve this question. \\
    - It's crucial to follow this structure and include both the chain of thought and the final answer in your response. \\
    \\
    \textcolor{blue}{Question}\\ 
    \texttt{<question>} \\
    The following translations from German to English contain a particular error. That error will be one of the following types: Named Entities: An entity (names, places, locations, etc.) is changed to a different entity. Numerical Values: Numerical values (ordinals or cardinals), dates, and/or units are changed. Modifiers or Adjectives: The modifiers and adjectives pertaining to a noun are changed. Negation or Antonyms: Introduce or remove a negation or change comparatives to their antonyms. Facts: Trivial factual errors not pertaining to the above classes are introduced in the translations. Dropped Content: A significant clause in the translation is removed. Please identify that error.  Source: Die Mexikanische Königsnatter oder San-Luis-Potosi-Königsnatter ist eine Schlange aus der Familie der Nattern. \\
    Translation: The Mexican King's Snake or San Luis Potosi King snake is a snake of the Lantern family. \\
    The translation contains an error pertaining to \\
    Options: \\
    (A) Modifiers or Adjectives \\
    (B) Numerical Values \\
    (C) Negation or Antonyms \\
    (D) Named Entities \\
    (E) Dropped Content \\
    (F) Facts \\
    \texttt{</question>} \\[1ex]
    \\
    \textcolor{blue}{Model Response}\\  
    \texttt{<chain\_of\_thought>} \\
    To identify the error in the translation, let's compare the source sentence with the translated sentence step by step. \\
    1. \textbf{Named Entities}: The source sentence mentions "Die Mexikanische Königsnatter oder San-Luis-Potosi-Königsnatter" which translates to "The Mexican King's Snake or San Luis Potosi King snake." The names of the snake species seem to be correctly translated, although the formatting and capitalization might slightly differ, they refer to the same entities. \\
    2. \textbf{Numerical Values}: There are no numerical values mentioned in either the source or the translated sentence, so there's no error related to numerical values. \\
    3. \textbf{Modifiers or Adjectives}: The source sentence describes the snake as "eine Schlange aus der Familie der Nattern," which means "a snake from the family of colubrids" or "a snake of the family of snakes." The translation mentions "a snake of the Lantern family." Here, "Nattern" (which refers to a family of snakes, specifically colubrids) is changed to "Lantern family," which seems to be incorrect as "Lantern" does not correspond to any known family of snakes related to the context. \\
    4. \textbf{Negation or Antonyms}: There is no introduction or removal of negation, nor a change to comparatives or their antonyms in the translation. \\
    5. \textbf{Facts}: The change from "family of snakes (Nattern)" to "Lantern family" introduces a factual error since "Lantern family" does not correctly represent the family of snakes. \\
    6. \textbf{Dropped Content}: There doesn't seem to be a significant clause removed from the translation. \\
    Given these observations, the error seems to pertain to \textbf{Named Entities} or \textbf{Facts}. However, more precisely, it involves changing the family of the snake from "Nattern" (a correct term for a family of snakes) to "Lantern family," which is not a recognized family of snakes. This change affects the factual accuracy regarding the classification of the snake. \\
    Since "Lantern family" is not a correct family name and represents a change in factual information about the snake's family, and considering that "Named Entities" typically refers to names of specific entities like people, places, or organizations rather than general terms like families of animals, the error is best categorized under \textbf{Facts}. \\
    \texttt{</chain\_of\_thought>} \\

    \texttt{<final\_answer>} \\
    \textcolor{red}{F} \\
    \texttt{</final\_answer>} \\
    \hline
  \end{tabular}
  }
\caption{Example of DIP Prompt (N=1) and model response for translation error detection in the BBH-Induct dataset. The gold (correct) answer for this example is (D) Named Entities.}
\label{tab:DIPPromptN1TranslationError}
\end{figure*}

\begin{figure*}[ht!]
  \centering
  {\tiny  
  \begin{tabular}{p{0.97\linewidth}}
  \hline
    \textcolor{blue}{Zero-shot CoT Prompt} \\
    \texttt{<task\_instruction>} \\
    Detect the type of error in an English translation of a German source sentence. \\
    \texttt{</task\_instruction>} \\[0.5ex]
    \textbf{Note} Do not use programming or code to solve this question. \\[1ex]
    \\
    \textcolor{blue}{Prompt Template} \\
    \texttt{<output\_format>} \\
    The output is a single letter corresponding to the correct option from the list (e.g., A, B, C, etc.). \\
    \texttt{</output\_format>} \\

    Follow these steps carefully: \\
    1. Provide a step-by-step chain of thought that answers the question. \\
    \quad \texttt{<chain\_of\_thought>} \\
    \quad [Your step-by-step chain of thought here] \\
    \quad \texttt{</chain\_of\_thought>} \\

    2. Based on your \texttt{chain\_of\_thought}, provide the final answer according to the rules specified in the \texttt{output\_format} section. \\
    If unsure and "Output Format" is option, guess the closest option. Present your final answer in the following format: \\
    \quad \texttt{<final\_answer>} \\
    \quad [Your final answer here] \\
    \quad \texttt{</final\_answer>} \\[1ex]
    \\
    \textcolor{blue}{Question}\\ 
    \texttt{<question>} \\
    The following translations from German to English contain a particular error. That error will be one of the following types: Named Entities: An entity (names, places, locations, etc.) is changed to a different entity. Numerical Values: Numerical values (ordinals or cardinals), dates, and/or units are changed. Modifiers or Adjectives: The modifiers and adjectives pertaining to a noun are changed. Negation or Antonyms: Introduce or remove a negation or change comparatives to their antonyms. Facts: Trivial factual errors not pertaining to the above classes are introduced in the translations. Dropped Content: A significant clause in the translation is removed. Please identify that error.  Source: Die Mexikanische Königsnatter oder San-Luis-Potosi-Königsnatter ist eine Schlange aus der Familie der Nattern. \\
    Translation: The Mexican King's Snake or San Luis Potosi King snake is a snake of the Lantern family. \\
    The translation contains an error pertaining to \\
    Options: \\
    (A) Modifiers or Adjectives \\
    (B) Numerical Values \\
    (C) Negation or Antonyms \\
    (D) Named Entities \\
    (E) Dropped Content \\
    (F) Facts \\
    \texttt{</question>} \\[1ex]
    \\
    \textcolor{blue}{Model Response}\\  
    \texttt{<chain\_of\_thought>} \\
    \#\# Step 1: Identify the source sentence and its translation. \\
    The source sentence in German is: "Die Mexikanische Königsnatter oder San-Luis-Potosi-Königsnatter ist eine Schlange aus der Familie der Nattern." \\
    The English translation provided is: "The Mexican King's Snake or San Luis Potosi King snake is a snake of the Lantern family." \\
    \#\# 2: Compare the source sentence and its translation to identify potential errors. \\
    - The source sentence mentions "Mexikanische Königsnatter" and "San-Luis-Potosi-Königsnatter," which are translated to "Mexican King's Snake" and "San Luis Potosi King snake," respectively. This seems accurate in terms of named entities. \\
    - The source sentence states the snake is "eine Schlange aus der Familie der Nattern," which translates to "a snake from the family of Nattern" or more accurately in English, "a snake from the family of colubrids" (since Nattern refers to a family of non-venomous snakes, commonly known as colubrids). \\
    - The translation mentions "the Lantern family," which does not match "der Familie der Nattern" (the family of colubrids or non-venomous snakes). \\
    \#\# 3: Determine the type of error. \\
    - \textbf{Named Entities (D):} The names of the snakes ("Mexican King's Snake" and "San Luis Potosi King snake") seem to be correct translations of "Mexikanische Königsnatter" and "San-Luis-Potosi-Königsnatter." \\
    - \textbf{Numerical Values (B):} There are no numerical values mentioned in the source sentence or the translation. \\
    - \textbf{Modifiers or Adjectives (A):} The modifiers and adjectives seem to be generally preserved, though the critical error lies elsewhere. \\
    - \textbf{Negation or Antonyms (C):} There is no introduction or removal of negation or change of comparatives to their antonyms. \\
    - \textbf{Facts (F):} The family mentioned in the translation ("Lantern family") is incorrect. The correct family should be related to "Nattern," which refers to colubrids, not "Lantern," which could imply a different or fictional classification. \\
    - \textbf{Dropped Content (E):} The translation does not seem to have dropped a significant clause but has introduced an incorrect fact. \\
    \#\# 4: Conclusion on the type of error. \\
    The error is in the family of the snake, which is translated as "the Lantern family" instead of something akin to "the family of colubrids" or "the family of non-venomous snakes." This change introduces a \textbf{factual error} regarding the classification of the snake. \\
    \#\# 5: Select the correct option based on the analysis. \\
    Given that the error involves changing a factual piece of information about the snake's family from "Nattern" (colubrids) to "Lantern family," this falls under \textbf{Facts (F)}: Trivial factual errors not pertaining to the above classes are introduced in the translations. \\
    \texttt{</chain\_of\_thought>} \\

    The final answer is: \\
    \texttt{<final\_answer>} \\
    \textcolor{red}{F} \\
    \texttt{</final\_answer>} \\
    \hline
  \end{tabular}
  }
\caption{Example of Zero-shot CoT Prompt and model response for translation error detection in the BBH-Induct dataset. The gold (correct) answer for this example is (D) Named Entities.}
\label{tab:ZeroShotCoTTranslationError}
\end{figure*}

\begin{figure*}[ht!]
  \centering
  {\tiny  
  \begin{tabular}{p{0.97\linewidth}}
  \hline
    \textcolor{blue}{S-CoT Prompt} \\
    \texttt{<task\_instruction>} \\
    Detect the type of error in an English translation of a German source sentence. \\
    \texttt{</task\_instruction>} \\[1ex]
    \\
    \textcolor{blue}{Prompt Template} \\
    \texttt{<output\_format>} \\
    The output is a single letter corresponding to the correct option from the list (e.g., A, B, C, etc.). \\
    \texttt{</output\_format>} \\

    Follow these steps carefully: \\
    1. Carefully consider the problem and generate the strategic knowledge that would best guide the problem-solving process. \\
    \quad \texttt{<strategy>} \\
    \quad [Your strategy here] \\
    \quad \texttt{</strategy>} \\

    2. Provide a step-by-step chain of thought that answers the question. \\
    \quad \texttt{<chain\_of\_thought>} \\
    \quad [Your step-by-step chain of thought here] \\
    \quad \texttt{</chain\_of\_thought>} \\

    3. Based on your \texttt{chain\_of\_thought}, provide the final answer according to the rules specified in the \texttt{output\_format} section. \\
    If unsure and "Output Format" is option, guess the closest option. Present your final answer in the following format: \\
    \quad \texttt{<final\_answer>} \\
    \quad [Your final answer here] \\
    \quad \texttt{</final\_answer>} \\

    \textbf{Note} Do not use programming or code to solve this question. \\
    \\
    \textcolor{blue}{Question}\\ 
    \texttt{<question>} \\
    The following translations from German to English contain a particular error. That error will be one of the following types: Named Entities: An entity (names, places, locations, etc.) is changed to a different entity. Numerical Values: Numerical values (ordinals or cardinals), dates, and/or units are changed. Modifiers or Adjectives: The modifiers and adjectives pertaining to a noun are changed. Negation or Antonyms: Introduce or remove a negation or change comparatives to their antonyms. Facts: Trivial factual errors not pertaining to the above classes are introduced in the translations. Dropped Content: A significant clause in the translation is removed. Please identify that error.  Source: Die Mexikanische Königsnatter oder San-Luis-Potosi-Königsnatter ist eine Schlange aus der Familie der Nattern. \\
    Translation: The Mexican King's Snake or San Luis Potosi King snake is a snake of the Lantern family. \\
    The translation contains an error pertaining to \\
    Options: \\
    (A) Modifiers or Adjectives \\
    (B) Numerical Values \\
    (C) Negation or Antonyms \\
    (D) Named Entities \\
    (E) Dropped Content \\
    (F) Facts \\
    \texttt{</question>} \\[1ex]
    \\
    \textcolor{blue}{Model Response}\\  
    \texttt{<strategy>} \\
    To identify the type of error in the translation, let's first analyze the source sentence and the translation given, and then compare them to determine the nature of the discrepancy. \\
    1. \textbf{Source Sentence}: Die Mexikanische Königsnatter oder San-Luis-Potosi-Königsnatter ist eine Schlange aus der Familie der Nattern. \\
    2. \textbf{Translation}: The Mexican King's Snake or San Luis Potosi King snake is a snake of the Lantern family. \\
    The key components to compare are: \\
    - Names of the species \\
    - Family of the species \\
    \texttt{</strategy>} \\

    \texttt{<chain\_of\_thought>} \\
    1. \textbf{Named Entities and Modifiers}: \\
    \quad - The source sentence mentions "Mexikanische Königsnatter" and "San-Luis-Potosi-Königsnatter" as the names of the snake, which are translated to "The Mexican King's Snake" and "San Luis Potosi King snake". The names are similar but not exactly the same due to the capitalization and slight variation in wording ("King's" vs. "King"), yet they refer to the same entities. \\
    2. \textbf{Family of the Species}: \\
    \quad - The source sentence states that the snake is "aus der Familie der Nattern", which translates to "from the family of colubrids" or more generally "from the family of snakes" (Nattern is a German term for a type of snake, specifically the Colubridae family). \\
    \quad - The translation states the snake is "of the Lantern family". \\
    3. \textbf{Error Identification}: \\
    \quad - The term "Nattern" refers to a specific family of snakes (Colubridae), which is not the "Lantern family". The "Lantern family" does not correspond to any known classification in biology related to snakes. This indicates a factual error regarding the family classification of the snake. \\
    4. \textbf{Type of Error}: \\
    \quad - \textbf{Named Entities (D)}: The names are slightly adjusted but still refer to the same entities. \\
    \quad - \textbf{Numerical Values (B)}: There are no numerical values mentioned. \\
    \quad - \textbf{Modifiers or Adjectives (A)}: The modifiers are adjusted but not in a way that changes the meaning significantly regarding the error type. \\
    \quad - \textbf{Negation or Antonyms (C)}: There is no introduction or removal of negation. \\
    \quad - \textbf{Facts (F)}: The family of the snake is incorrectly stated as "Lantern family" instead of a correct biological classification like "Colubridae" or a similar term. \\
    \quad - \textbf{Dropped Content (E)}: No significant clause is removed. \\
    Given the analysis, the error pertains to \textbf{Facts (F)} because a trivial factual error is introduced by stating the snake belongs to the "Lantern family", which is not a correct biological classification. \\
    \texttt{</chain\_of\_thought>} \\

    \texttt{<final\_answer>} \\
    \textcolor{red}{F} \\
    \texttt{</final\_answer>} \\
    \hline
  \end{tabular}
  }
\caption{Example of S-CoT Prompt and model response for translation error detection in the BBH-Induct dataset. The gold (correct) answer for this example is (D) Named Entities.}
\label{tab:SCOTTranslationError}
\end{figure*}

\begin{figure*}[ht!]
  \centering
  {\tiny  
  \begin{tabular}{p{0.97\linewidth}}
  \hline
    \textcolor{blue}{R-CoT Prompt} \\
    \texttt{<task\_instruction>} \\
    Detect the type of error in an English translation of a German source sentence. \\
    \texttt{</task\_instruction>} \\[1ex]
    \\
    \textcolor{blue}{Prompt Template} \\
    \texttt{<output\_format>} \\
    The output is a single letter corresponding to the correct option from the list (e.g., A, B, C, etc.). \\
    \texttt{</output\_format>} \\

    Follow these steps carefully: \\
    1. First, carefully consider the question and propose a clear reasoning framework to guide your thinking. \\
    \quad \texttt{<reasoning\_framework>} \\
    \quad Step 1: [First high-level reasoning step] \\
    \quad Step 2: [Second high-level reasoning step] \\
    \quad Step 3: [Third high-level reasoning step] \\
    \quad [Add additional steps if necessary] \\
    \quad \texttt{</reasoning\_framework>} \\

    2. Next, use the framework you've outlined to explain your reasoning step-by-step. \\
    \quad \texttt{<chain\_of\_thought>} \\
    \quad 1. [Explanation aligned with Step 1] \\
    \quad 2. [Explanation aligned with Step 2] \\
    \quad 3. [Explanation aligned with Step 3] \\
    \quad [Continue as necessary] \\
    \quad \texttt{</chain\_of\_thought>} \\

    3. Based on your \texttt{chain\_of\_thought}, provide the final answer according to the rules specified in the \texttt{output\_format} section. \\
    If unsure and "Output Format" is option, guess the closest option. Present your final answer in the following format: \\
    \quad \texttt{<final\_answer>} \\
    \quad [Your final answer here] \\
    \quad \texttt{</final\_answer>} \\

    \textbf{Note:} Do not use programming or code to solve this question. \\
    \\
    \textcolor{blue}{Question}\\ 
    \texttt{<question>} \\
    The following translations from German to English contain a particular error. That error will be one of the following types: Named Entities: An entity (names, places, locations, etc.) is changed to a different entity. Numerical Values: Numerical values (ordinals or cardinals), dates, and/or units are changed. Modifiers or Adjectives: The modifiers and adjectives pertaining to a noun are changed. Negation or Antonyms: Introduce or remove a negation or change comparatives to their antonyms. Facts: Trivial factual errors not pertaining to the above classes are introduced in the translations. Dropped Content: A significant clause in the translation is removed. Please identify that error.  Source: Die Mexikanische Königsnatter oder San-Luis-Potosi-Königsnatter ist eine Schlange aus der Familie der Nattern. \\
    Translation: The Mexican King's Snake or San Luis Potosi King snake is a snake of the Lantern family. \\
    The translation contains an error pertaining to \\
    Options: \\
    (A) Modifiers or Adjectives \\
    (B) Numerical Values \\
    (C) Negation or Antonyms \\
    (D) Named Entities \\
    (E) Dropped Content \\
    (F) Facts \\
    \texttt{</question>} \\[1ex]
    \\
    \textcolor{blue}{Model Response}\\  
    \texttt{<reasoning\_framework>} \\
    Step 1: Identify the key elements in the source sentence and the translation. \\
    Step 2: Compare the key elements to determine the type of error. \\
    Step 3: Analyze the differences and match them to the given error types. \\
    \texttt{</reasoning\_framework>} \\

    \texttt{<chain\_of\_thought>} \\
    1. \textbf{Identify key elements in the source sentence and the translation:} \\
    \quad - Source sentence: "Die Mexikanische Königsnatter oder San-Luis-Potosi-Königsnatter ist eine Schlange aus der Familie der Nattern." \\
    \quad - Translation: "The Mexican King's Snake or San Luis Potosi King snake is a snake of the Lantern family." \\
    \quad - Key elements include names (Mexikanische Königsnatter, San-Luis-Potosi-Königsnatter), classification (Familie der Nattern), and the English translation provided. \\
    2. \textbf{Compare the key elements:} \\
    \quad - The names "Mexikanische Königsnatter" and "San-Luis-Potosi-Königsnatter" are translated to "Mexican King's Snake" and "San Luis Potosi King snake," respectively. This seems accurate in terms of named entities. \\
    \quad - The family "der Nattern" is translated to "the Lantern family." \\
    3. \textbf{Analyze the differences and match them to the given error types:} \\
    \quad - The source sentence mentions "Familie der Nattern," which translates to "family of snakes" or more specifically "Colubridae" in scientific terms. \\
    \quad - The translation mentions "the Lantern family," which is incorrect because "Nattern" refers to a family of non-venomous snakes (Colubridae), not "Lantern." \\
    4. \textbf{Determine the error type:} \\
    \quad - The error involves changing "Nattern" (a family of snakes) to "Lantern family," which is a factual error regarding the classification of the snake. \\
    \quad - This does not fit into named entities (D) as the issue is not with the names but with the classification. \\
    \quad - It does not fit into numerical values (B), negation or antonyms (C), or modifiers/adjectives (A) directly. \\
    \quad - It does not fit into dropped content (E) as there is no significant clause removed. \\
    \quad - The closest match is \textbf{Facts (F)}, as it involves introducing a trivial factual error regarding the family classification. \\
    \texttt{</chain\_of\_thought>} \\

    \texttt{<final\_answer>} \\
    \textcolor{red}{F} \\
    \texttt{</final\_answer>} \\
    \hline
  \end{tabular}
  }
\caption{Example of R-CoT Prompt and model response for translation error detection in the BBH-Induct dataset.  The gold (correct) answer for this example is (D) Named Entities.}
\label{tab:RCOTTranslationError}
\end{figure*}

\end{document}